\pdfoutput=1
\documentclass[11pt,dvipsnames]{article}

\usepackage{acl}

\usepackage{times}
\usepackage{latexsym}
\usepackage{lipsum}
\usepackage{tabularx}
\usepackage{amsfonts}
\usepackage{amssymb}
\usepackage{verbatim}
\usepackage{makecell}

\usepackage[T1]{fontenc}
\usepackage[T5]{fontenc}
\usepackage[utf8]{inputenc}

\usepackage{microtype}

\usepackage{inconsolata}

\DeclareMathAlphabet{\mathbbold}{U}{bbold}{m}{n}

\usepackage{graphicx}
\usepackage{multicol}
\usepackage{multirow}
\usepackage{booktabs}
\usepackage{amsmath}
\usepackage{upgreek}
\usepackage{footmisc}
\usepackage{tablefootnote}
\usepackage{footnote}
\usepackage{enumitem}
\usepackage{xcolor}
\usepackage{tcolorbox}
\usepackage{pifont}
\usepackage{subcaption} % For subfigures
\usepackage{worldflags}
\usepackage{xurl}

\newcommand{\formatnumber}[1]{%
  \ifnum#1<1000
    #1
  \else
    \ifnum#1<1000000
      \fpeval{round(#1/1000,1)}k
    \else
      \fpeval{round(#1/1000000,1)}M
    \fi
  \fi
}

\newcommand{\mybar}[3]{%
  \formatnumber{#1}%
  & 
  \begin{minipage}{100pt} % Ensures bars are in one line
    {\color{Green!80}%
      \rule{\fpeval{(#2/#1)*75}pt}{8pt}%
    }%
    {\color{Gray!50}%
      \rule{\fpeval{(1 - #2/#1)*75}pt}{8pt}%
    } \\
    \centering {\small{\fpeval{round((#2/#1)*100,2)}\%}}
    \ifx#3\relax % If #3 is empty, do nothing
    \else
      \textsuperscript{#3} % Adds superscript (for "*", it will display "*")
    \fi
  \end{minipage}%
}

\newcommand{\tagbox}[2]{%
  \tcbox[colback=#1!20,colframe=#1,arc=5pt,boxrule=0.8pt,left=2pt,right=2pt,top=1pt,bottom=1pt, on line]{\small{#2}}%
}

\newcommand{\cmark}{\tcbox[colback=Green!30,colframe=Green,arc=6pt,boxrule=0pt,left=1pt,right=1pt,top=1pt,bottom=1pt, on line]{\ding{51}}} % ✔
\newcommand{\xmark}{\tcbox[colback=Gray!20,colframe=Gray,arc=6pt,boxrule=0pt,left=1pt,right=1pt,top=1pt,bottom=1pt, on line]{\ding{55}}} % ✖

\newcommand{\increase}[1]{\tcbox[colback=Green!15,colframe=Green,arc=6pt,boxrule=0pt,left=0.2pt,right=0.2pt,top=0.2pt,bottom=0.2pt, on line]{\small(+#1)}}
\newcommand{\decrease}[1]{\tcbox[colback=BrickRed!15,colframe=BrickRed,arc=6pt,boxrule=0pt,left=0.2pt,right=0.2pt,top=0.2pt,bottom=0.2pt, on line]{\small(-#1)}}
\newcommand{\nochange}[1]{\tcbox[colback=Gray!15,colframe=Gray,arc=6pt,boxrule=0pt,left=0.2pt,right=0.2pt,top=0.2pt,bottom=0.2pt, on line]{\small(0.00)}}

\setlist{topsep=1pt,itemsep=1pt,partopsep=1pt, parsep=1pt}

\title{Crowdsource, Crawl, or Generate? Creating SEA-VL,\\a Multicultural Vision-Language Dataset for Southeast Asia}

\usepackage{skak}

\def\quad{\hskip0.75em\relax}

\author{
\normalsize{\textbf{Samuel Cahyawijaya$^{\symqueen, 1, 2, 3}$}}\quad
\normalsize{\textbf{Holy Lovenia$^{\symqueen, 2, 3}$}}\quad
\normalsize{\textbf{Joel Ruben Antony Moniz$^{\symqueen, 4, 5}$}}\quad
\normalsize{\textbf{Tack Hwa Wong$^{\symqueen, 6}$}}\quad
\\
\normalsize{\textbf{Mohammad Rifqi Farhansyah$^{\symknight, 7}$}}\quad
\normalsize{\textbf{Thant Thiri Maung$^{\symknight, 8}$}}\quad
\normalsize{\textbf{Frederikus Hudi$^{\symknight, 9, 10, 2}$}}\quad
\normalsize{\textbf{David Anugraha$^{\symknight, 11}$}}\quad
\\
\normalsize{\textbf{Muhammad Ravi Shulthan Habibi$^{\symknight, 12, 2, 3}$}}\quad
\normalsize{\textbf{Muhammad Reza Qorib$^{\symknight, 13}$}}\quad
\normalsize{\textbf{Amit Agarwal$^{\symknight, 14}$}}\quad
\\
\normalsize{\textbf{Joseph Marvin Imperial$^{\symknight, 15, 16}$}}\quad
\normalsize{\textbf{Hitesh Laxmichand Patel$^{\symknight, 14}$}}\quad
\normalsize{\textbf{Vicky Feliren$^{\symknight, 17}$}}\quad
\normalsize{\textbf{Bahrul Ilmi Nasution$^{\symknight, 18}$}}\quad
\\
\normalsize{\textbf{Manuel Antonio Rufino$^{\symknight, 19}$}}\quad
\normalsize{\textbf{Genta Indra Winata$^{\symknight, 20, 2, 3}$}}\quad
\normalsize{\textbf{Rian Adam Rajagede$^{\symknight, 21}$}}\quad
\normalsize{\textbf{Carlos Rafael Catalan$^{\symknight, 19}$}}\quad
\\
\normalsize{\textbf{Mohamed Fazli Imam$^{22}$}}\quad
\normalsize{\textbf{Priyaranjan Pattnayak$^{6}$}}\quad
\normalsize{\textbf{Salsabila Zahirah Pranida$^{22}$}}\quad
\normalsize{\textbf{Kevin Pratama$^{23}$}}\quad
\\
\normalsize{\textbf{Yeshil Bangera$^{24}$}}\quad
\normalsize{\textbf{Adisai Na-Thalang$^{25}$}}\quad
\normalsize{\textbf{Patricia Nicole Monderin$^{19}$}}\quad
\normalsize{\textbf{Yueqi Song$^{26}$}}\quad
\normalsize{\textbf{Christian Simon$^{27}$}}\quad
\\
\normalsize{\textbf{Lynnette Hui Xian Ng$^{26}$}}\quad
\normalsize{\textbf{Richardy Lobo' Sapan$^{12}$}}\quad
\normalsize{\textbf{Taki Hasan Rafi$^{28}$}}\quad
\normalsize{\textbf{Bin Wang$^{29}$}}\quad
\normalsize{\textbf{Supryadi$^{30}$}}\quad
\\
\normalsize{\textbf{Kanyakorn Veerakanjana$^{31}$}}\quad
\normalsize{\textbf{Piyalitt Ittichaiwong$^{31}$}}\quad
\normalsize{\textbf{Matthew Theodore Roque$^{19}$}}\quad
\normalsize{\textbf{Karissa Vincentio$^{3, 32}$}}\quad
\\
\normalsize{\textbf{Takdanai Kreangphet$^{33}$}}\quad
\normalsize{\textbf{Phakphum Artkaew$^{34}$}}\quad
\normalsize{\textbf{Kadek Hendrawan Palgunadi$^{35}$}}\quad
\normalsize{\textbf{Yanzhi Yu}$^{36}$}\quad
\\
\normalsize{\textbf{Rochana Prih Hastuti$^{37}$}}\quad
\normalsize{\textbf{William Nixon$^{7}$}}\quad
\normalsize{\textbf{Mithil Bangera$^{24}$}}\quad
\normalsize{\textbf{Adrian Xuan Wei Lim$^{13}$}}\quad
\\
\normalsize{\textbf{Aye Hninn Khine$^{38}$}}\quad
\normalsize{\textbf{Hanif Muhammad Zhafran$^{7}$}}\quad
\normalsize{\textbf{Teddy Ferdinan$^{39}$}}\quad
\normalsize{\textbf{Audra Aurora Izzani$^{40}$}}\quad
\\
\normalsize{\textbf{Ayushman Singh$^{20}$}}\quad
\normalsize{\textbf{Evan$^{6}$}}\quad
\normalsize{\textbf{Jauza Akbar Krito$^{6}$}}\quad
\normalsize{\textbf{Michael Anugraha$^{6}$}}\quad
\normalsize{\textbf{Fenal Ashokbhai Ilasariya$^{6}$}}\quad
\\
\normalsize{\textbf{Haochen Li$^{6}$}}\quad
\normalsize{\textbf{John Amadeo Daniswara$^{6}$}}\quad
\normalsize{\textbf{Filbert Aurelian Tjiaranata$^{12}$}}\quad
\normalsize{\textbf{Eryawan Presma Yulianrifat$^{12}$}}\quad
\\
\normalsize{\textbf{Can Udomcharoenchaikit$^{41}$}}\quad
\normalsize{\textbf{Fadil Risdian Ansori$^{6}$}}\quad
\normalsize{\textbf{Mahardika Krisna Ihsani$^{22}$}}\quad
\normalsize{\textbf{Giang Nguyen$^{42}$}}\quad
\\
\normalsize{\textbf{Anab Maulana Barik$^{13}$}}\quad
\normalsize{\textbf{Dan John Velasco$^{19}$}}\quad
\normalsize{\textbf{Rifo Ahmad Genadi$^{22}$}}\quad
\normalsize{\textbf{Saptarshi Saha$^{43}$}}\quad
\normalsize{\textbf{Chengwei Wei$^{29}$}}\quad
\\
\normalsize{\textbf{Isaiah Flores$^{44}$}}\quad
\normalsize{\textbf{Kenneth Ko Han Chen$^{45}$}}\quad
\normalsize{\textbf{Anjela Gail Santos$^{46}$}}\quad
\normalsize{\textbf{Wan Shen Lim$^{26}$}}\quad
\normalsize{\textbf{Kaung Si Phyo$^{45}$}}\quad
\\
\normalsize{\textbf{Tim Santos$^{47}$}}\quad
\normalsize{\textbf{Meisyarah Dwiastuti$^{48}$}}\quad
\normalsize{\textbf{Jiayun Luo$^{6}$}}\quad
\normalsize{\textbf{Jan Christian Blaise Cruz$^{22, 2}$}}\quad
\normalsize{\textbf{Ming Shan Hee$^{49}$}}\quad
\\
\normalsize{\textbf{Ikhlasul Akmal Hanif$^{12}$}}\quad
\normalsize{\textbf{M.Alif Al Hakim$^{12}$}}\quad
\normalsize{\textbf{Muhammad Rizky Sya'ban$^{7}$}}\quad
\normalsize{\textbf{Kun Kerdthaisong$^{50}$}}\quad
\\
\normalsize{\textbf{Lester James V. Miranda$^{51}$}}\quad
\normalsize{\textbf{Fajri Koto$^{22, 2, 3}$}}\quad
\normalsize{\textbf{Tirana Noor Fatyanosa$^{52}$}}\quad
\normalsize{\textbf{Alham Fikri Aji$^{22, 2, 3}$}}\quad
\\
\normalsize{\textbf{Jostin Jerico Rosal$^{53}$}}\quad
\normalsize{\textbf{Jun Kevin$^{54}$}}\quad
\normalsize{\textbf{Robert Wijaya$^{\symknight, 49}$}}\quad
\normalsize{\textbf{Onno P. Kampman$^{\symknight, 55, 2}$}}\quad
\\
\normalsize{\textbf{Ruochen Zhang$^{\symknight, 56, 2}$}}\quad
\normalsize{\textbf{Börje F. Karlsson$^{\symknight, 57}$}}\quad
\normalsize{\textbf{Peerat Limkonchotiwat$^{\symknight, 58, 59, 2}$}}\quad
\\
\\
\parbox{\textwidth}{\centering
\small{$^{1}$Cohere}\quad
\small{$^{2}$SEACrowd}\quad
\small{$^{3}$IndoNLP}\quad
\small{$^{4}$Mila~-~Quebec~AI~Institute}\quad
\small{$^{5}$Polytechnique~Montreal}\quad
\small{$^{6}$Independent}\quad
\small{$^{7}$Bandung~Institute~of~Technology}\quad
\small{$^{8}$Ton~Duc~Thang~University}\quad
\small{$^{9}$Nara~Institute~of~Science~and~Technology}\quad
\small{$^{10}$Works~Applications}\quad
\small{$^{11}$University~of~Toronto}\quad
\small{$^{12}$University~of~Indonesia}\quad
\small{$^{13}$National~University~of~Singapore}\quad
\small{$^{14}$Oracle}\quad
\small{$^{15}$University~of~Bath}\quad
\small{$^{16}$National~University~Philippines}\quad
\small{$^{17}$Monash~University,~Indonesia}\quad
\small{$^{18}$The~University~of~Manchester}\quad
\small{$^{19}$Samsung~R\&D~Institute~Philippines}\quad
\small{$^{20}$Capital~One}\quad
\small{$^{21}$Universitas~Islam~Indonesia}\quad
\small{$^{22}$MBZUAI}\quad
\small{$^{23}$Meta}\quad
\small{$^{24}$University~of~New~Haven}\quad
\small{$^{25}$SCB~10X}\quad
\small{$^{26}$Carnegie~Mellon~University}\quad
\small{$^{27}$Sony~Group~Corporation}\quad
\small{$^{28}$Hanyang~University}\quad
\small{$^{29}$Institute~for~Infocomm~Research,~Singapore}\quad
\small{$^{30}$Tianjin~University}\quad
\small{$^{31}$Faculty~of~Medicine~Siriraj~Hospital,~Mahidol~University}\quad
\small{$^{32}$Binus~University}\quad
\small{$^{33}$Srinakharinwirot~University}\quad
\small{$^{34}$New~York~University}\quad
\small{$^{35}$Institut~Teknologi~Sepuluh~Nopember}\quad
\small{$^{36}$Macau~University~of~Science~and~Technology}\quad
\small{$^{37}$Universitas~Gadjah~Mada}\quad
\small{$^{38}$King~Mongkut's~University~of~Technology~Thonburi}\quad
\small{$^{39}$Wrocław~Tech}\quad
\small{$^{40}$University~of~Illiinois,~Urbana-Champaign}\quad
\small{$^{41}$Vidyasirimedhi~Institute~of~Science~and~Technology}\quad
\small{$^{42}$Auburn~University}\quad
\small{$^{43}$Indian~Statistical~Institute,~Kolkata}\quad
\small{$^{44}$Ateneo~de~Manila~University}\quad
\small{$^{45}$Singapore~Polytechnic}\quad
\small{$^{46}$University~of~the~Philippines}\quad
\small{$^{47}$Graphcore}\quad
\small{$^{48}$Dataxet:Sonar}\quad
\small{$^{49}$Singapore~University~of~Technology~and~Design}\quad
\small{$^{50}$Thammasat~University}\quad
\small{$^{51}$Allen~AI}\quad
\small{$^{52}$Brawijaya~University}\quad
\small{$^{53}$Seoul~National~University~of~Science~and~Technology}\quad
\small{$^{54}$Universitas~Pelita~Harapan}\quad
\small{$^{55}$MOH~Office~for~Healthcare~Transformation}\quad
\small{$^{56}$Brown~University}\quad
\small{$^{57}$Beijing~Academy~of~Artificial~Intelligence~(BAAI)}\quad
\small{$^{58}$AI~Singapore}\quad
\small{$^{59}$Chulalongkorn~University}\quad
}
\\
\small{\textbf{$^{\symqueen}$Main contributors}}\quad
\small{\textbf{$^{\symknight}$Major contributors}}
\\
}

\begin{document}
\maketitle
\begin{abstract}

Southeast Asia (SEA) is a region of extraordinary linguistic and cultural diversity, yet it remains significantly underrepresented in vision-language (VL) research. 
This often results in artificial intelligence (AI) models that fail to capture SEA cultural nuances. 
To fill this gap, we present \textbf{\href{https://huggingface.co/collections/SEACrowd/sea-vl-multicultural-vl-dataset-for-southeast-asia-67cf223d0c341d4ba2b236e7}{SEA-VL}}, an open-source initiative dedicated to developing high-quality, culturally relevant data for SEA languages. 
By involving contributors from SEA countries, SEA-VL aims to ensure better cultural relevance and diversity, fostering greater inclusivity of underrepresented languages in VL research.
Beyond crowdsourcing, our initiative goes one step further in the exploration of the automatic collection of culturally relevant images through crawling and image generation. 
First, we find that image crawling 
achieves approximately $\sim$85\% cultural relevance while being more cost- and time-efficient than crowdsourcing.
Second, despite the substantial progress in generative vision models, synthetic images remain unreliable in accurately reflecting SEA cultures.
The generated images often fail to reflect the nuanced traditions
and cultural contexts of the region. Collectively, we gather 1.28M SEA culturally-relevant images, more than 50 times larger than other existing datasets.
Through SEA-VL, we aim to bridge the representation gap in SEA, fostering the development of more inclusive AI systems that authentically represent diverse cultures across SEA.

% Our findings suggest the promising potential of collecting image data through image crawling which is able to reach $\sim$85\% cultural relevance while producing a larger amount of data within the comparable cost and time budgets compared to crowdsourcing. While image generation, despite the astonishing progress, is unreliable for creating images that accurately capture the cultural aspects in SEA. Moreover, licensing issues of current image generation models further hinder the applicability of image generation as a scalable automatic image collection method.

% This paper details the initial phase of the SEA-VL initiative, focusing on the development of reliable, scalable, and robust approaches for collecting culturally relevant images with descriptive annotations. 

\end{abstract}

\section{Introduction}

The rapid evolution of artificial intelligence (AI) and machine learning (ML) has produced increasingly sophisticated models capable of integrating textual and visual information. However, these advancements often disproportionately benefit certain languages and cultures~\cite{yong-etal-2023-prompting,pham2023semi,cahyawijaya-etal-2023-instructalign,10.1093/pnasnexus/pgae346,cahyawijaya-etal-2024-llms,myung2024blend,li2024culturellm,winata2024worldcuisines}, leaving underrepresented cultures---particularly those of Southeast Asia (SEA)---largely overlooked~\cite{aji2022one,winata-etal-2023-nusax,purwarianti-etal-2025-nusadialogue,cahyawijaya-etal-2024-cendol,urailertprasert-etal-2024-sea}. 
This disparity creates a significant challenge in developing AI technologies that effectively cater to the diverse cultural contexts of underrepresented regions.

Home to over 1,300 languages and rich cultural diversity, SEA is among the world's most linguistically vibrant regions~\cite{enfield2011linguistic,aji-etal-2022-one,lovenia-etal-2024-seacrowd}.
However, the lack of SEA-relevant datasets, particularly in the vision-language (VL) domain~\cite{lovenia-etal-2024-seacrowd}, limits AI accessibility and risks  cultural irrelevance or bias against SEA populations~\cite{winata2024worldcuisines,urailertprasert-etal-2024-sea,cahyawijaya2024llmeveryonerepresentingunderrepresented}.
Addressing this disparity by creating datasets that authentically capture SEA's linguistic and cultural nuances requires large-scale collaborative efforts~\cite{bell2021perspectives}.
Building on crowdsourcing initiatives like NusaCrowd~\cite{cahyawijaya-etal-2023-nusacrowd}, SEACrowd~\cite{lovenia-etal-2024-seacrowd}, and Aya Dataset~\cite{singh-etal-2024-aya}, SEA-VL takes a holistic approach to bridging the resource gap for SEA cultural representation in VL research.
Unlike existing efforts, which primarily focus on text-based tasks or limited subsets of visual data, SEA-VL aims to develop comprehensive, high-quality VL datasets that reflect SEA's cultural heritage and linguistic diversity.
SEA-VL seeks to address linguistic underrepresentation, trust, and dignity in AI, ensuring technological advancements benefit the diverse communities of SEA.

\begin{figure}
    \centering
    \includegraphics[width=\linewidth, trim={0 4.5cm 10cm 0}, clip]{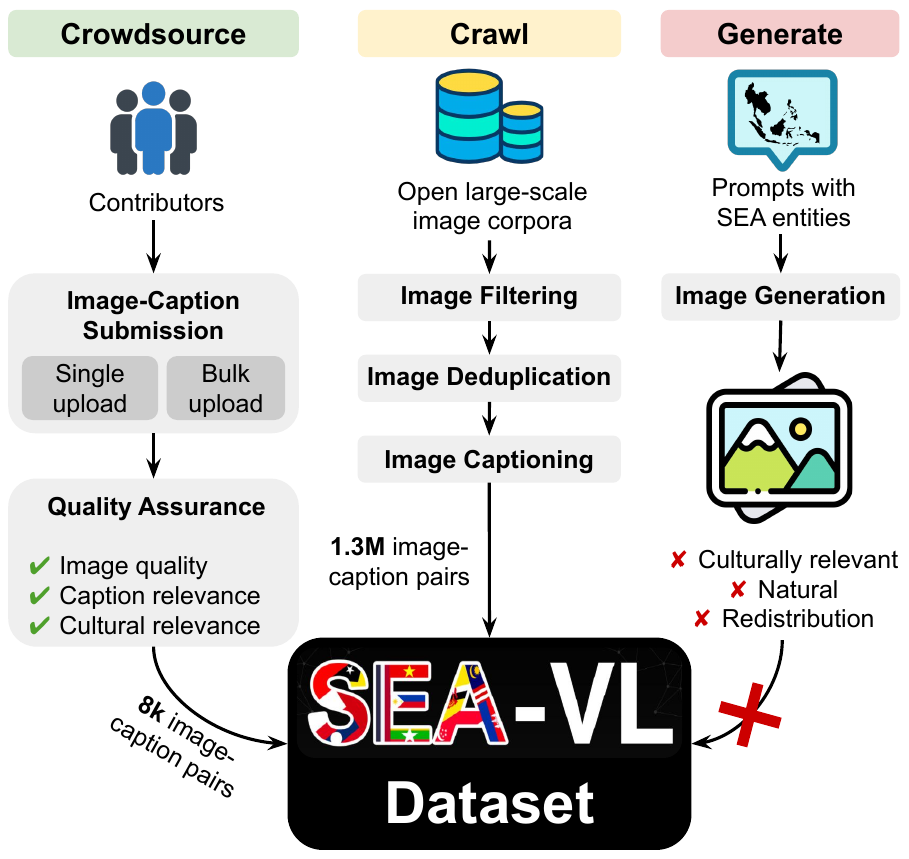}
    \caption{SEA-VL addresses the underrepresentation of SEA languages in vision-language research through a multipronged strategy for collecting culturally relevant images, incorporating image crowdsourcing, crawling, and synthetic generation.}
    \label{fig:enter-label}
    \vspace{-10pt}
\end{figure}

SEA-VL\footnote{SEA-VL dataset: \url{https://huggingface.co/collections/SEACrowd/sea-vl-multicultural-vl-dataset-for-southeast-asia-67cf223d0c341d4ba2b236e7}.} distinguishes itself from other grassroots community-driven initiatives by going beyond relying on manual data collection only, through the participation of local contributors.
With the recent popularity of various strong AI models, SEA-VL explores diverse methodologies to collect culturally relevant images in SEA.
Specifically, SEA-VL collects culturally relevant image data using three different methods: (1) manual human collection, (2) a validated data filtering and deduplication pipeline on crawled images, and (3) image generation through diffusion models.
To ensure that the data collected authentically represent the lived experiences and cultural contexts of the region, an extensive human evaluation by native participants is performed using different image collection methodologies.
This evaluation not only enhances the quality and relevance of the datasets, but also provides a better understanding of the feasibility, efficiency, and quality of using AI-based data collection solutions to produce culturally relevant VL datasets, specifically for the SEA region.
We also compare manual vs. automatic metadata collection to assess how well AI-based solutions generate valid, relevant metadata, which is beneficial for creating culturally relevant VL datasets.

\begin{table*}[t!]
\centering
\resizebox{\linewidth}{!}{
    \begin{tabular}{m{3.5cm}rcm{7cm}ccccccccccc}
    \toprule
    \textbf{Dataset Name} & \textbf{\#Images} & \textbf{\%SEA Images} & \textbf{Cultural Coverage$^\dagger$} & \textbf{ID} & \textbf{TH} & \textbf{PH} & \textbf{SG} & \textbf{MY} & \textbf{MM} & \textbf{BN} & \textbf{KH} & \textbf{LA} & \textbf{VN} & \textbf{TL} \\
    \midrule
    SEA-VQA~\small{\cite{urailertprasert-etal-2024-sea}} & \mybar{488}{488}{} & \tagbox{red}{Tradition \& Art} \tagbox{yellow}{Landmark} & \cmark & \cmark & \cmark & \cmark & \cmark & \xmark & \xmark & \cmark & \cmark & \cmark & \xmark \\
    \midrule
    WorldCuisines~\small{\cite{winata2024worldcuisines}} & \mybar{6000}{930}{} & \tagbox{orange}{Cuisine} & \cmark & \cmark & \cmark & \cmark & \cmark & \cmark & \cmark & \cmark & \cmark & \cmark & \xmark \\
    \midrule
    CVQA~\small{\cite{mogrovejo2024cvqa}} & \mybar{7000}{1200}{} & \tagbox{SkyBlue}{Daily Life} \tagbox{CarnationPink}{Local Products} \tagbox{Purple}{Pop Culture} \tagbox{yellow}{Landmark} \tagbox{red}{Tradition \& Art} \tagbox{Brown}{Transportation} \tagbox{Green}{Plant \& Animal} \tagbox{RoyalBlue}{Sport \& Recreation} \tagbox{orange}{Cuisine} & \cmark & \xmark & \cmark & \cmark & \xmark & \xmark & \xmark & \xmark & \xmark & \xmark & \xmark \\
    \midrule
    TotalDefMeme~\small{\cite{prakash2023totaldefmeme}} & \mybar{7200}{7200}{} & \tagbox{red}{Tradition \& Art} \tagbox{Purple}{Pop Culture} & \xmark & \xmark & \xmark & \cmark & \xmark & \xmark & \xmark & \xmark & \xmark & \xmark & \xmark \\
    \midrule
    OpenViVQA~\small{\cite{nguyen2023openvivqa}} & \mybar{11200}{11200}{} & Unknown & \xmark & \xmark & \xmark & \xmark & \xmark & \xmark & \xmark & \xmark & \xmark & \cmark & \xmark \\
    \midrule
    Bloom Library~\small{\cite{leong-etal-2022-bloom}} & \mybar{112000}{23000}{} & Unknown & \cmark & \cmark & \cmark & \cmark & \cmark & \cmark & \cmark & \cmark & \cmark & \cmark & \cmark \\
    \midrule
    CC3M~\small{\cite{sharma2018conceptual}} & \mybar{3020000}{3600}{*} & Unknown & \multicolumn{11}{c}{Unknown} \\
    \midrule
    WiT~\small{\cite{srinivasan2021wit}} & \mybar{11500000}{5750}{*} & Unknown & \cmark & \cmark & \cmark & \cmark & \cmark & \cmark & \cmark & \xmark & \xmark & \cmark & \xmark \\
    \bottomrule
    \textbf{SEA-VL} (ours) & \mybar{1280000}{1024000}{*} & \tagbox{SkyBlue}{Daily Life} \tagbox{CarnationPink}{Local Products} \tagbox{Purple}{Pop Culture} \tagbox{yellow}{Landmark} \tagbox{red}{Tradition \& Art} \tagbox{Brown}{Transportation} \tagbox{Green}{Plant \& Animal} \tagbox{RoyalBlue}{Sport \& Recreation} \tagbox{orange}{Cuisine} & \cmark & \cmark & \cmark & \cmark & \cmark & \cmark & \cmark & \cmark & \cmark & \cmark & \cmark \\
    \bottomrule
    \end{tabular}
}
\caption{Summary of potential datasets with culturally-relevant images, showing cultural and regional coverage. A checkmark {\tiny\cmark\hspace{1pt}} indicates coverage in the respective country (Appendix~\ref{app:sea-countries}), while a cross {\tiny\xmark\hspace{1pt}} indicates no coverage. SEA-VL has $>$50$\times$ SEA images compared to other existing datasets. $^\dagger$We follow the cultural category from \citet{mogrovejo2024cvqa}. $^*$The number is estimated based on cultural relevance in our human evaluation.}
\vspace{-6pt}
\label{tab:seavl-competitors}
\end{table*}

\noindent The contributions of SEA-VL are three-fold:
\begin{itemize}
    \item \textbf{Comprehensive, Culturally-Relevant VL Datasets:} SEA-VL develops high-quality, culturally rich VL datasets that reflect SEA's linguistic and cultural diversity. By actively engaging local contributors, SEA-VL ensures that the data authentically represents lived experiences and regional contexts.
    \item \textbf{Analysis of Trade-offs in Data Collection Methods:} SEA-VL analyzes the trade-offs between effectiveness and efficiency across data collection methodologies, and demonstrates the strengths and weaknesses when employing different strategies.
    \item \textbf{Assessment of AI-based Solutions:} SEA-VL assesses the feasibility along with the efficiency and quality of AI-driven methods for collecting image data and its metadata, comparing these methods with manual processes for creating regionally relevant VL datasets.
    % \item \textbf{Inclusive AI Development:} By providing accessible and high-quality resources, SEA-VL supports the creation of AI systems that are equitable, culturally aware, and linguistically inclusive for Southeast Asian populations.
\end{itemize}

\section{Related Work}

\paragraph{Crowdsource-based Data Collection}
% \todo{Genta, Blaise, Aji, Fajri, Bahrul}
% \dummy{\lipsum[1]}

Crowdsourcing is historically widely used in the machine learning community as a means to collect large amounts of high-quality human data \cite{Crescenzi2017}. 
Compared to alternatives such as scraping \cite{taesiri2024imagenet}, crowdsourcing's main advantage is the ability to explicitly highlight granular variables such as demographics, opinions, and regional variations \cite{mostafazadeh-davani-etal-2024-d3code}.
The increased interest in the development of multilingual large language models (LLMs) in recent years has pushed crowdsourcing as a powerful strategy for grassroots-led data collection \cite{lovenia-etal-2024-seacrowd,naggita10.1145/3600211.3604659} where representation is a key factor. As LLM research continues to grow, culture-grounded benchmarks have begun to gain traction as strong challengers for models that culturally lean towards the West \cite{mogrovejo2024cvqa,winata2024worldcuisines,taesiri2024imagenet}. Such benchmarks are reliant on crowdsourcing to achieve the breadth and granularity needed to accurately portray multiculturality.

\paragraph{Underrepresented Cultures across the World}
% \todo{Genta, Blaise, Aji, Fajri, Bahrul}
% \dummy{\lipsum[1]}

Efforts for improving tools, models, and resources for low-resource languages have increased in recent years partly driven by underrepresentation in widely-adopted LLMs and benchmark datasets \cite{pham2023semi,globalbench,khanuja-etal-2024-image,urailertprasert-etal-2024-sea}. Beyond Southeast Asia, grassroots-led organizations have successfully spearheaded efforts to produce resources for underrepresented cultures in their region.

Masakhane, a grassroots group based in Africa, has produced work that contributes strong benchmarks \cite{adelani-etal-2021-masakhaner,adelani-etal-2022-masakhaner,adelani-etal-2023-masakhanews}, models~\cite{dossou-etal-2022-afrolm} and evaluation metrics~\cite{wang-etal-2024-afrimte} to alleviate resource scarcity and assess the direct applicability of widely used benchmarking methods towards non-English languages. Similarly, AI4Bharat has developed an extensive body of work, including benchmarks~\cite{verma2025milumultitaskindiclanguage}, datasets~\cite{jain2024bhasaanuvaadspeechtranslationdataset}, tools~\cite{khan-etal-2024-indicllmsuite,ai4bharat2024indicvoicesr}, and models~\cite{gala2024airavataintroducinghindiinstructiontuned} representing the Indian subcontinent. Significant efforts have also been made to promote languages indigenous to the Americas, spearheaded by the AmericasNLP community. Notable projects include textual inference, such as AmericaNLI~\cite{ebrahimi2022americasnli}, as well as advancements in machine translation~\cite{mager2023neural,rangel2024advancing}. These groups also host workshops and shared tasks~\cite{adelani-etal-2022-findings} to promote research interest.
% in their languages. 

Beyond region-wide representation, recent work has also begun to pay attention to granularity \textit{within} countries. Resources such as MC$^2$ \cite{zhang-etal-2024-mc2} and CultureAtlas \cite{fung2024massively} produce benchmarks that highlight differing cultural variations practiced within one country, further emphasizing the issue of representing a country with only one cultural norm. There is also a strong emphasis on dialectal research, with groups like ACL SIGARAB advocating for the inclusion of diverse Arabic dialects—each with distinct morphological and stylistic variations—whereas benchmarks often rely solely on \textit{Modern Standard} Arabic to represent the entire Middle East and North Africa region \cite{abdul-mageed-etal-2024-nadi}. While interest in underrepresented languages and cultures has grown in recent years, there remains a significant gap compared to the prevalence of English in models and datasets. Efforts on Southeast Asian cultures, in particular, still need improvement in areas such as multimodality---a research gap that we strive to overcome through SEA-VL and other related open community initiatives.

\section{Image Collection in SEA-VL}

The goal of SEA-VL is to improve the representation of SEA cultures in VL research through various image collection strategies and to provide in-depth assessments on the trade-off of each strategy. SEA-VL employs three strategies: image crowdsourcing, image crawling, and image generation. In addition, SEA-VL explores methods to gather metadata from the collected images automatically.

% \begin{figure}[!t]
%     \centering
%     \includegraphics[width=0.65\linewidth]{images/image_deduplication.png}
%     \caption{Deduplication}
%     \label{fig:enter-label}
% \end{figure*}

\subsection{Image Crowdsourcing}
% \todo{Holy}

Despite the potentially higher noise, crowdsourcing has become a common strategy employed as a means for large-scale data collection~\cite{cahyawijaya-etal-2023-nusacrowd,lovenia-etal-2024-seacrowd,singh-etal-2024-aya}. Prior works have shown that improving the data quality through data pruning brings substantial benefits to the model capability~\cite{marion2023less,chen2024alpagasus,longpre2024pretrainers, singh-etal-2024-aya}. To further improve the quality and cultural relevance of the collected images to the SEA context, we also conduct a quality assurance phase to curate and gather feedback.

\paragraph{Image Collection}
For image collection, we ask contributors to submit only those images they personally own, avoiding images retrieved from publicly accessible platforms. Contributors upload their images through a designated form, providing metadata on the location where the image was taken and to which of the 11 SEA countries (Appendix~\ref{app:sea-countries}) it is relevant to. In addition, they indicate their native language and are required to include a caption in both English and their native language. Submission guidelines also specify that images must be culturally relevant, and any personally identifiable information (PII), such as faces and license plates, must be redacted before submission.

\paragraph{Quality Assurance}

After data collection, we conduct a quality check, where at least two people validate each image. If the inter-annotator agreement between two validators on a certain image is below 80\%, we add another annotator for that image. Contributors must pass a screening test before participating in quality assurance. Validators determine whether an image meets quality standards, assess its cultural relevance on a 5-point scale, and verify the appropriateness of its caption. Appendix~\ref{app:image-crowdsourcing-qa} presents more details about QA.
% as reported in Table X.

\subsection{Image Crawling} \label{sec:collection_crawling}
% \todo{Samuel}
Despite the rise of crowdsourced data collection, many efforts are actively managed only in their early stages, with enthusiasm waning over time, posing sustainability and scalability challenges~\cite{cahyawijaya-etal-2023-nusacrowd,lovenia-etal-2024-seacrowd,gehrmann-etal-2022-gemv2,singh-etal-2024-aya}. To address this, SEA-VL explores autonomous methods to gather culturally relevant images in SEA by crawling existing sources. A carefully designed pipeline ensures high-quality collection through curated filtering and deduplication.

\paragraph{Image Filtering}
% \todo{Samuel, Peerat, Reza, Hudi}

The goal of image filtering is to select SEA culturally-relevant images from a large set of images. Based on our assessment of various image filtering strategies (see Appendix~\ref{app:image-filtering-method}), we perform image filtering through semantic similarity. Given a set of unfiltered images $I_{uf}$, we filter out images that have an average semantic similarity score lower than a threshold $\rho$ compared with a set of reference culturally-relevant images $I_{ref}$. Specifically, given an input image $x \in I_{uf}$, we define the image filtering function $f(x)$ as:
\begin{align}
    f(x) = \mathbbold{1}_{[\frac{1}{|I_{\mathrm{ref}}|} \sum\limits_{z \in I_\mathrm{ref}}  \Psi(\lambda(x), \lambda(z)) ] \geq \rho},
    \label{eq:image-filtering}
\end{align}
% \begin{align} \mathrm{score}(x) &= \frac{1}{\lvert I_{\mathrm{ref}} \rvert} \sum_{z \in I_{\mathrm{ref}}} \Psi\bigl(\lambda(x),\lambda(z)\bigr), \\ f(x) &= \begin{cases} 1, & \text{if } \mathrm{score}(x) \ge \rho,\\ 0, & \text{otherwise}. \end{cases} \label{eq:image-filtering} \end{align}
where $\mathbbold{1}$ denotes an indicator function, $\lambda$ denotes an image encoding function, and $\Psi$ denotes a cosine similarity function between two image representations. Given $\Psi$, $\lambda$, and $I_\mathrm{ref}$, we tune the value of $\rho$ to ensure that we end up with high-quality, culturally relevant images after filtering.

% \begin{table*}[!t]
%     \centering
%     \resizebox{\linewidth}{!}{
%         \begin{tabular}{c|c|c|c|c|c|c|c|c|c}
%              \toprule
%              \multirow{2}{*}{\textbf{Threshold}} & \multicolumn{3}{c|}{\textbf{CC3M}} & \multicolumn{3}{c|}{\textbf{COYO}} & \multicolumn{3}{c}{\textbf{WiT}} \\
%              \cmidrule(lr){2-4} \cmidrule(lr){5-7} \cmidrule(lr){8-10}
%              & \textbf{Relevance} & \textbf{\#Images} & \textbf{\%Images} & \textbf{Relevance} & \textbf{\#Images} & \textbf{\%Images}  & \textbf{Relevance} & \textbf{\#Images} & \textbf{\%Images}  \\
%              \midrule
%              $[51.5\dots52.5)$ & 58\% & 11885 & 0.40\% & 54\% & 8925 & 0.54\% & 80\% & 9627 & 0.23\% \\
%              $[52.5\dots53.5)$ & 70\% & 6824 & 0.23\% & 40\% & 5919 & 0.36\% & 82\% & 6715 & 0.13\% \\
%              $[53.5\dots54.5)$ & 78\% & 3841 & 0.13\% & 70\% & 3996 & 0.24\% & 92\% & 4377 & 0.06\% \\
%              $[54.5\dots55.5)$ & 84\% & 2091 & 0.07\% & 82\% & 2323 & 0.14\% & 94\% & 2590 & 0.03\% \\
%              $\geq55.5$ & 92\% & 1499 & 0.05\% & 78\% & 2294 & 0.14\% & 90\% & 2162 & 0.02\% \\
%              \bottomrule
%         \end{tabular}
%     }
%     \caption{The human evaluation result of the image filtering phase on CC3M, COYO, and WiT  datasets.}
%     \label{tab:image_filtering}
% \end{table*}

\begin{figure*}
    \centering
    \resizebox{\linewidth}{!}{
        \includegraphics{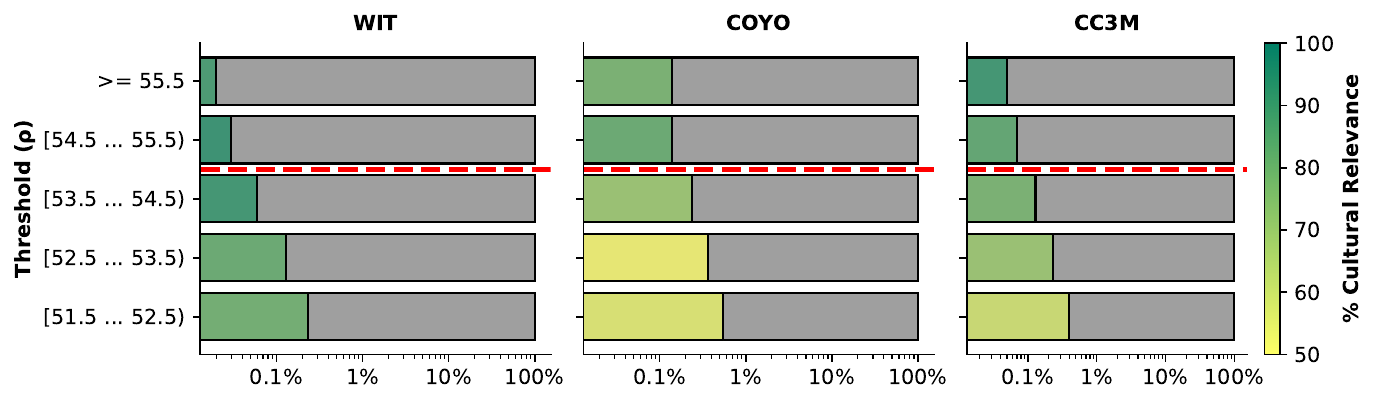}
    }
    \caption{Human evaluation results of SEA image filtering on CC3M, COYO, and WiT datasets. Grey area indicates the proportion of images below the similarity threshold ($\rho$). We take the top-2 threshold groups ($\geq$54.5) as the final threshold retaining $\sim$0.15\% of the total images with $\sim$85\% cultural relevance.}
    \label{fig:image_filtering}
\end{figure*}

\paragraph{Image Deduplication}
% \todo{Samuel, Carlos, Feliren}
% \dummy{\lipsum[1]}

Collecting data by crawling various sources tends to result in highly duplicated collections~\cite{sharma2018conceptual,kakaobrain2022coyo-700m}, causing a skewed representation towards certain image concepts. Mitigating this problem, we incorporate an effective and efficient image deduplication process after filtering the images. Image deduplication can be thought of as an unsupervised image clustering problem, where a pair of images that are closely similar is considered redundant. Specifically, given two images $x, y \in I_{uf}$ and a minimum threshold $\epsilon$, we define an image deduplication function $g(x, y)$ as:
\begin{align}
    g(x, y) = \mathbbold{1}_{\Psi(\lambda(x), \lambda(y)) < \epsilon},
\end{align}
where $\mathbbold{1}$ denotes an indicator function, $\lambda$ denotes an image encoding function, and $\Psi$ denotes a similarity function between two image representations. We explore two groups of methods for image deduplication, i.e., perceptual hashing~\cite{hadmi2012perceptual,hamadouche2021comparative} and semantic similarity~\cite{wang2014learning,radford2021learning}. Perceptual hashing encodes an image into a binary hash code, while semantic similarity encodes an image into a normalized real-valued vector.

\subsection{Image Generation}
% \todo{Samuel, Anton, Wong}
With the rise of various diffusion-based image generation models~\cite{sohldickstein2015deepunsupervisedlearningusing} such as Stable Diffusion~\cite{rombach2022highresolutionimagesynthesislatent,esser2024scalingrectifiedflowtransformers} and DALL-E~\cite{pmlr-v139-ramesh21a,betker2023improving}, we further explore the possibility of generating SEA culturally relevant synthetic images. In principle, the inference of a diffusion model reverses the diffusion process by gradually transforming random noise to obtain a sample from the desired distribution. This process is repeated several steps, gradually refining the sample until it resembles the desired distribution, resulting in samples that are diverse and realistic. On the other hand, recently proposed autoregressive image generation models~\cite{chameleonteam2024chameleonmixedmodalearlyfusionfoundation,sun2024autoregressivemodelbeatsdiffusion,wu2024janusdecouplingvisualencoding} also show promising image generation quality; unlike diffusion-based models, these models generate images in an autoregressive manner using discrete image tokens.
% similar to LLMs. 
% Autoregressive image generation models encode an image using an image encoder such as CLIP~\cite{radford2021learning} and SigLIP~\cite{zhai2023sigmoid}, and transform them into discrete token representation for autoregressive generation~\cite{van2017neural,esser2021taming,sun2024autoregressive}.

\subsection{Image Captioning} \label{sec:img_captioning}
% \todo{Samuel, Wong, Fazli}
In order to make the automatically collected images more meaningful, we conducted several attempts to infer image metadata, such as captions. We originally intended to explore image captioning in both English and the target language of the respective SEA culture; however, due to the poor quality of captioning in the target language (as shown in Appendix~\ref{app:image-caption-failure}), we narrow our attempt to focus on prompting for generating culturally relevant captions in English. 
% The image categories that we infer follow the cultural categories in CVQA~\cite{mogrovejo2024cvqa}.

\section{Experiment Details}

\paragraph{Image Filtering}
% \todo{Samuel, Peerat, Carlos, Anton, Reza}
We incorporate image semantic similarity for our image filtering pipeline.\footnote{We explore various strategies for image filtering such as heuristics filtering based on metadata and image-text similarity (see Appendix~\ref{app:image-filtering-method}).} To determine the optimal threshold $\rho$ for collecting culturally relevant SEA images, we conduct human evaluations on 3 datasets: Conceptual Captions (CC3M)~\cite{sharma2018conceptual}, COYO~\cite{kakaobrain2022coyo-700m}, and WiT~\cite{srinivasan2021wit}. We use the SEA region images of CVQA~\cite{mogrovejo2024cvqa} and all of SEA-VQA~\cite{urailertprasert-etal-2024-sea} as the reference images $I_\mathrm{ref}$. Through an exploratory data analysis, we drop all images with a similarity score below 0.515, as only a tiny fraction of images below that threshold range are culturally relevant. This process removes $\sim$99\% of all the images in the datasets. We cluster the remaining images into 5 groups, each with a different threshold range. We then randomly sample 50 images from each group and conduct a human evaluation to measure the cultural relevance of each group (see Appendix~\ref{app:eval-filtering}).

\paragraph{Image Deduplication}
% \todo{Samuel, Carlos, Feliren}
For image perceptual hashing, we utilize the implementation from pHash~\cite{zauner2010implementation}, which encodes an image into a 64-bit binary hash code and then uses the Hamming distance as a measure of similarity. For semantic similarity, we use 3 different image embedding models, i.e., CLIP-ViT (86M)~\cite{radford2021learning}, SigLIP (878M)~\cite{zhai2023sigmoid}, and Nomic Embed Vision v1.5 (92M)~\cite{nussbaum2024nomic}. We perform image deduplication on the images collected from our image filtering experiment and crowdsourcing. The embedding models encode an image into a normalized embedding vector, after which cosine similarity is computed between two images. Then, we perform a human evaluation, taking 50 pairs of the top predicted samples of each method and evaluating their correctness using the criteria defined in Appendix~\ref{app:eval-duplication}. We use 1 RTX3050 for all embedding-based models and CPU for perceptual hashing.

\paragraph{Image Generation}
% \todo{Samuel, Peerat, Carlos, Anton, Reza}
We evaluate three diffusion-based image generation models: Stable Diffusion 2~\cite{Rombach_2022_CVPR}, Stable Diffusion 3.5~\cite{esser2024scalingrectifiedflowtransformers} and FLUX.1-dev~\cite{flux2023}, and one autoregressive model: Janus-Pro (7B)~\cite{chen2025janusprounifiedmultimodalunderstanding}. Images are generated for 3 cultural aspects: food, landmarks, and traditions.

For food, images are generated using the prompt template ``An image of people eating X'' where X is the name of a Southeast Asian dish based on a list derived from the WorldCuisines dataset~\cite{winata2024worldcuisines}. For landmarks, the prompt is ``An image of people at X'', where X is a UNESCO World Heritage Site~\cite{unesco_whc} in Southeast Asia. For traditions, we use the prompt ``An image of people doing X'', where X is the name of a UNESCO Intangible Cultural Heritage retrieved from the metadata of SEA-VQA\footnote{The metadata for SEA-VQA is not publicly released and was obtained directly from the paper's authors.}~\cite{urailertprasert-etal-2024-sea}.

We report the detailed hyperparameters used for each model in Appendix~\ref{app:hyperparameters}. To evaluate the quality, we sample 50 generated images from each category and manually inspect them by comparing them with images from the crowdsourcing and crawling stages on two aspects: correctness and naturalness (see Appendix~\ref{app:eval-image-gen} for details).

\paragraph{Image Captioning}

For image captioning, we explore 4 multilingual vision-language models (VLMs) within our experiments: Qwen2-VL (7B)~\cite{Qwen-VL,Qwen2VL}, Pangea (7B)~\cite{yue2024pangeafullyopenmultilingual}, 
PaliGemma2 (10B)~\cite{steiner2024paligemma2familyversatile}, and
Maya (8B)~\cite{alam2024mayainstructionfinetunedmultilingual}. To find the best way to collect culturally-relevant image captions, we conduct an evaluation on 2 prompting methods, i.e., location-agnostic and location-aware promptings~\cite{mogrovejo2024cvqa}. We prompt all image captioning models to highlight these cultural items, such as local food, traditions, landmarks, or other relevant elements. The prompt should be concise, consisting of 3 to 5 sentences. The specific prompts and hyperparameters are detailed in Appendix~\ref{app:hyperparameters}. We manually inspect 50 random caption generations per method. See Appendix~\ref{app:eval-captioning} for more evaluation details. 
% For image categorization, we utilize the same similarity function as defined in Equation~\ref{eq:image-filtering} but use the reference images $I^\mathrm{ref}$ from all the CVQA images grouped by their category. For each image, we then assign the category based on the group with the highest similarity score with the corresponding image. We evaluate the correctness of the category with human evaluation in Appendix~\ref{app:eval-labeling}.

\section{Results and Analysis}

\begin{table}[!t]
    \centering
    \resizebox{\linewidth}{!}{
        \begin{tabular}{l|c|c|c}
             \toprule
             \textbf{Model} & \textbf{\#Param} & \textbf{Precision} & \textbf{Throughput} \\
             \midrule
                Perceptual Hashing & - & 2.00\% & 48.72  \\
                \midrule
                CLIP-ViT & 86M & 32.67\% & 20.34 \\
                Nomic Embed Vis. & 92M & \textit{48.67\%} & 21.73 \\
                SigLIP (SO) & 400M & \textbf{59.33\%} & 3.91  \\
             \bottomrule
        \end{tabular}
    }
    \caption{Human evaluation result of the image deduplication over 50 top predicted samples. Throughput refers to the number of images processed per second.}
    \label{tab:image_deduplication}
    \vspace{-10pt}
\end{table}

\subsection{Image Filtering}

The human evaluation results of image filtering with different threshold ranges are shown in Figure~\ref{fig:image_filtering}. To ensure high cultural relevance, we select the two highest threshold groups ($\geq$54.5) for our image filtering pipeline. Using this threshold, we reach $\sim$85\% cultural-relevance with inter-annotator agreement ($\gamma$ coefficient) of 0.6410 while retaining only $\sim$0.1\% of the total images from the original dataset, e.g., from 3M images in CC3M, we gather 3,590. Using this curated threshold value, we scale the process of image filtering up to the full set of LAION~\cite{schuhmann2021laion} and COYO~\cite{kakaobrain2022coyo-700m}, with a total of $\sim$1.28B images\footnote{We collected 2.1B image URLs, but only $\sim$60\% of the images can be downloaded, on account of outdated links.}. From these two sources, we gather $\sim$1.72M SEA culturally-relevant images.
% \footnote{Out of 1.1B image URLS, we can only download 60\% of the images since many of the links are already outdated.}. 
We show the image distribution per dataset in Appendix~\ref{app:image-filtering-result}

\subsection{Image Deduplication}

As shown in Table~\ref{tab:image_deduplication}, perceptual hashing yields a very low score compared to all semantic-similarity-based methods.
% and is only able to correctly predict duplicated images that are almost exactly identical to the source image
This demonstrates the benefits of using pre-trained vision models and VLMs to extract semantic features from images. Among different pre-trained embedding models used, SigLIP shows the best performance in identifying duplicate images, with a 59.33\% precision score, compared to CLIP-ViT and Nomic Embed Vision achieving 32.67\% and 48.67\%, respectively. This demonstrates that scaling models improves scene identification. Despite the higher precision, the substantially larger number of parameters of SigLIP results in much lower inference throughput. In the case of large-scale image deduplication, smaller yet performant alternative such as Nomic Embed Vision is a more suitable option as it maximizes the throughput while retaining a high deduplication precision. We then run our deduplication pipeline using Nomic Embed Vision on the $\sim$1.72M images collected from LAION and COYO resulting in $\sim$1.27M  unique culturally-relevant images.

\begin{figure}[t]
    \centering
    \resizebox{0.95\linewidth}{!}{
        \includegraphics[trim={0 0 0 0.25cm}, clip]{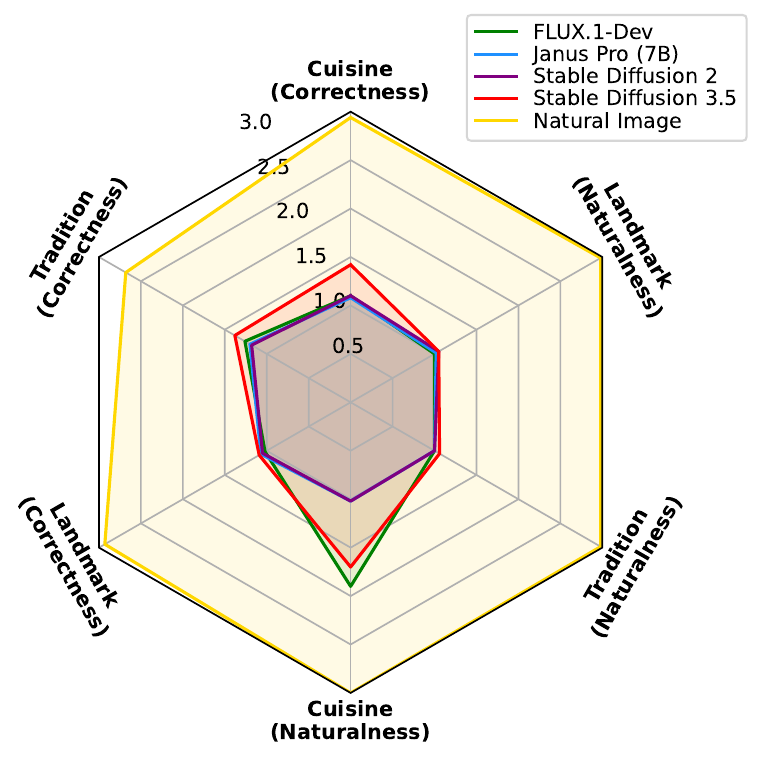}
    }
    \caption{Human evaluation result of SEA image generation with 3-point Likert score. Natural Image refers to non-generated images taken in real life.}
    \label{fig:image_generation}
    % \vspace{-10pt}
\end{figure}

\begin{table*}[!t]
    \centering
    \resizebox{0.9\linewidth}{!}{
        \begin{tabular}{lcccc}
            \toprule
            \multirow{2}{*}{\textbf{Model}} & \multicolumn{2}{c}{\textbf{SEA-VQA}} & \multicolumn{2}{c}{\textbf{WorldCuisines}} \\
            \cmidrule(lr){2-3} \cmidrule(lr){4-5}
            & \textbf{Correctness} & \textbf{Naturalness} & \textbf{Correctness} & \textbf{Naturalness} \\
            \toprule 
            Human & \textbf{2.68} & \textbf{2.82} & \textbf{2.98} & \textbf{2.96} \\
            \midrule
            MAYA (8B) & 1.62 \increase{0.26} & 2.34 \increase{0.14} & 2.20 \increase{0.02} & \underline{2.74} \nochange{0.00} \\
            PALI Gemma 2 (10B) & 2.04 \increase{0.06} & 1.72 \decrease{0.10} & 2.26 \decrease{0.16} & 1.74 \increase{0.04} \\
            Pangea (7B) & \underline{2.36} \decrease{0.24} & 2.42 \decrease{0.38} & \underline{2.48} \decrease{0.34} & 2.52 \decrease{0.16} \\
            Qwen2-VL (7B) & 2.10 \increase{0.36} & \underline{2.44} \nochange{0.00} & 2.24 \decrease{0.14} & 2.70 \decrease{0.36} \\
            \bottomrule
        \end{tabular}
    }
    \caption{Human evaluation result of the image captioning phase. We use a 3-point Likert score. The values in parentheses indicate the score shift
    (\tcbox[colback=Green!15,colframe=Green,arc=6pt,boxrule=0pt,left=0.2pt,right=0.2pt,top=0.2pt,bottom=0.2pt, on line]{\small{increase}}
    or
    \tcbox[colback=BrickRed!15,colframe=BrickRed,arc=6pt,boxrule=0pt,left=0.2pt,right=0.2pt,top=0.2pt,bottom=0.2pt, on line]{\small{decrease}})
     when incorporating location-aware prompting.
    }
    \label{tab:image_captioning}
    % \vspace{-8pt}
\end{table*}

\subsection{Image Generation}

The results in Figure~\ref{fig:image_generation} demonstrate that existing image generation models struggle to produce culturally relevant SEA images. Among the evaluated models, Stable Diffusion 3.5 yields the best performance, achieving the highest correctness scores of 1.42 and 1.38 for cuisine and tradition with a moderate naturalness rating of 1.70 for cuisine. However, image generation models still fall far short of human-collected images, which achieve near-perfect correctness and naturalness scores across all categories: correctness scores remain alarmingly low, with the best model scoring <1.5 in all categories; similarly, naturalness scores are notably poor, with all models producing highly unnatural images with scores barely exceeding 1.0. This highlights a critical gap of image generation models in capturing the essence of SEA cultural elements. 
% These highlight a critical gap of existing image generation models in generating synthetic image that captures the complexity and authenticity of SEA culture.

\subsection{Image Captioning}

% \paragraph{Image Captioning}

As shown in Table~\ref{tab:image_captioning}, existing VLMs can generate reasonably accurate and natural English captions for culturally relevant SEA images, though they still fall behind human-generated captions. Among the models tested, Pangea (7B) and Qwen2-VL (7B) performed best overall, with Pangea (7B) achieving the highest correctness scores in the location-agnostic setting. In comparison, Qwen2-VL (7B) excels in the location-aware setting. This suggests that existing VLMs can be a reliable option for generating synthetic captions in English. Nonetheless, there is still a huge gap for image captioning in local languages across SEA, as detailed in Appendix~\ref{app:image-caption-failure}. Our results also highlight that location-aware prompting does not consistently improve caption quality across models. For example, Qwen2-VL (7B) benefits from location-aware information in SEA-VQA but saw little improvement in WorldCuisines. Meanwhile, MAYA (8B) and PALI Gemma 2 (10B) show mixed results, with location-aware prompting slightly enhancing naturalness but not significantly improving correctness. Overall, while current models can generate high-quality English captions, there remains a gap between machine and human performance in terms of both cultural accuracy and linguistic fluency. This issue becomes more severe when the captions are in local languages, as described in Appendix~\ref{app:image-caption-failure}.

\begin{figure}[ht]
    \centering
    \begin{subfigure}[b]{\linewidth}
        \centering
        \includegraphics[width=\linewidth, trim={0 2.5cm 6.5cm 0}, clip]{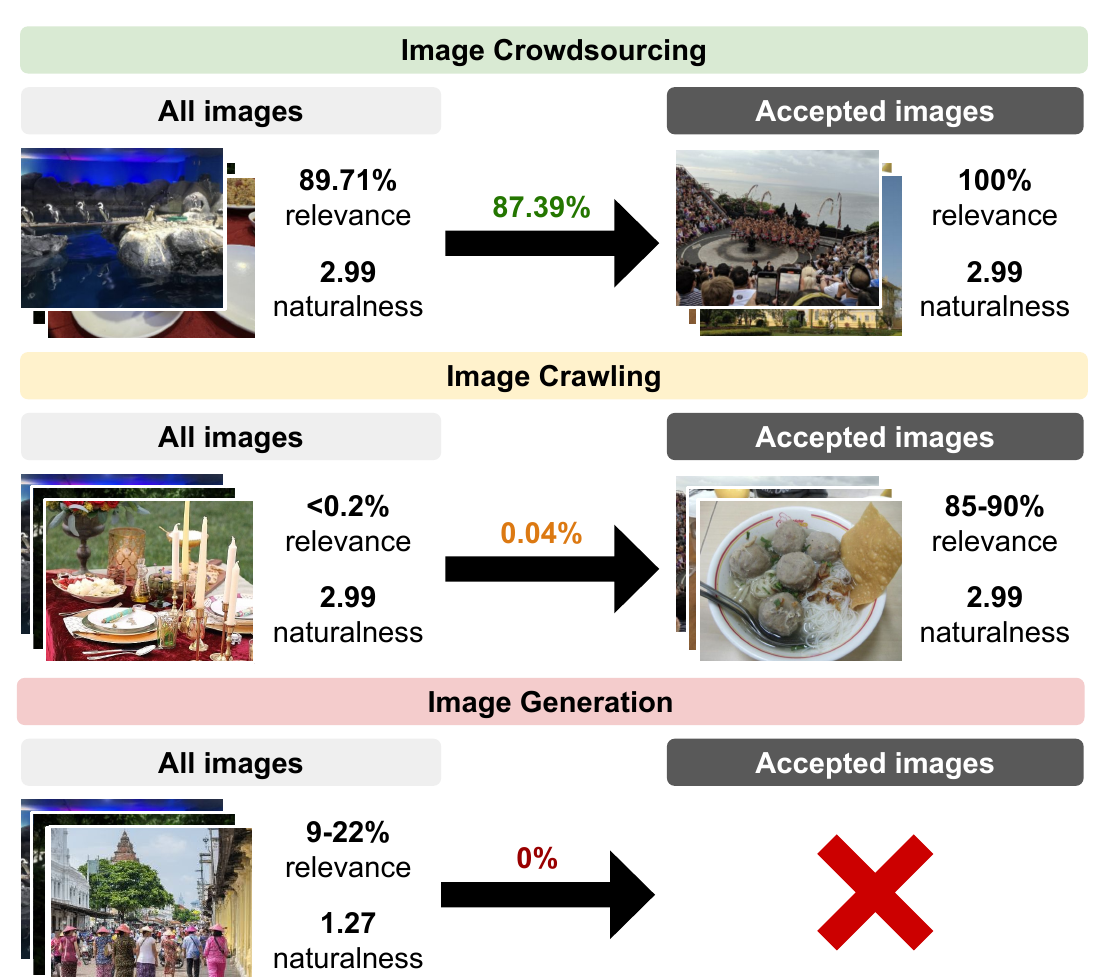}
        \caption{Image Quality}
        \label{fig:seavl_image_collection_image}
    \end{subfigure}
    \vspace{1em} % Optional space between subfigures
    \begin{subfigure}[b]{\linewidth}
        \centering
        \includegraphics[width=\linewidth, trim={0 13.5cm 9.5cm 0}, clip]{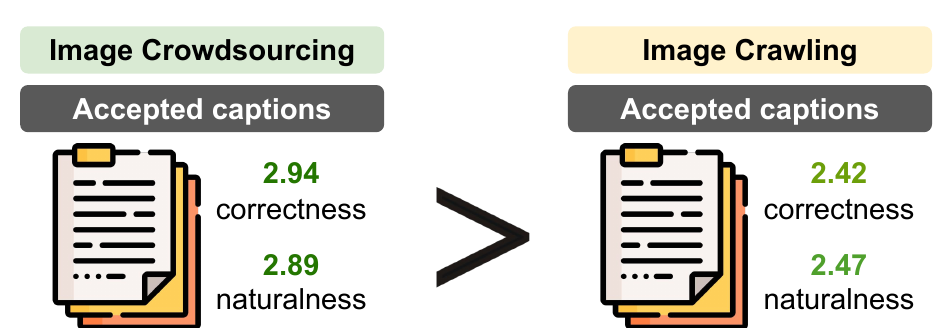}
        \caption{Caption Quality}
        \label{fig:seavl_image_collection_caption}
    \end{subfigure}
    \caption{Summary of different data collection strategies in SEA-VL. Relevance of image generation is from human evaluation on correctness. Relevance of image crowdsourcing is from annotation during quality assurance. Image naturalness is from naturalness evaluation of natural and generated images.}
    \label{fig:seavl_image_collection}
\end{figure}

% \paragraph{Image Labeling}
% \dummy{\lipsum[4]}

\section{Discussion}

% \begin{figure*}
%     \centering
%     \includegraphics[width=\linewidth]{images/human-in-the-loop-annotation.png}
%     \caption{Human-machine interaction enabled better model-based annotation increasing the annotation throughput and the scalability of the system over time.}
%     \label{fig:enter-label}
% \end{figure*}
% \dummy{\lipsum[3]}
% https://www.researchgate.net/figure/A-human-in-the-loop-approach-enables-scalable-pixel-level-annotation-of-large-image_fig1_349750818

\subsection{Resource Collected from SEA-VL}

By virtue of all the aforementioned data collection techniques, SEA-VL, at the time of writing, is the largest gathered cultural image database for SEA, with $\sim$1.28M culturally relevant images across SEA. This is more than 10$\times$ larger than existing works, as illustrated in Table~\ref{tab:seavl-competitors}. Specifically, SEA-VL collects 8k manually collected images from crowdsourcing and $\sim$495k automatically filtered crawled images.\footnote{We do not include the results from image generation in our published dataset due to licensing and the low cultural relevance of the generated images.} SEA-VL also brings broader outreach throughout SEA reaching underrepresented regions, e.g., Cambodia, Laos, and Timor Leste. Furthermore, SEA-VL also enables higher representation across different cultures in SEA as demonstrated by the high cultural coverage across all regions in SEA as exemplified in Table~\ref{tab:seavl-competitors} and Appendix~\ref{app:seavl_dist}.
With this extensive cultural and regional coverage of SEA-VL, we hope that AI models trained on SEA-VL can better understand and generate culturally accurate representations of SEA cultures, reducing biases and inaccuracies in representing SEA cultural contexts.

% \subsection{Crowdsource or Not Crowdsource?}
\subsection{Crowdsource, Crawl, or 
% Synthetically 
Generate?}
% \todo{Holy, Ruochen, Sam}
% (NB: Took a first pass here that hopefully makes it easier to fill this out, feel free to modify it as heavily as needed -- Joel)}
Figure \ref{fig:seavl_image_collection} shows a clear trade-off between crowdsourcing and filtering crawled images in existing corpora. While crowdsourcing produces exceptionally high-quality images, it requires significant effort. Over 85 days, we collected 10k crowdsourced images through a labor-intensive endeavor. However, the resulting images were extremely relevant ($\geq$89\%) and featured high caption quality (2.94). In contrast, filtering crawled images required only four days while still producing fairly high-quality results, with highly relevant images ($\geq$85\%) and a reasonably good caption quality (2.42).

In addition, as we hint at in Section~\ref{sec:collection_crawling}, crawling is more sustainable, with the potential to setup fully automated pipelines to continuously refresh data with minimal human intervention once initial filtering thresholds are established. In contrast to these methods, we find 
% synthetic 
image generation to be completely unviable as a data collection strategy, particularly on images that require cultural nuance, as in our case. Moreover, the current image generation models come with restrictive licenses, which further limit the feasibility of using image generation as a sustainable solution for an automated culturally-relevant image collection method.

Thus, our overall recommendation might be to rely on filtering crawled images for a scalable solution, such as for the creation of large-scale training sets; crowdsourcing, on the other hand, despite being effort-intensive, is extremely useful as a means of obtaining images of high quality (e.g., creating challenging test set). At the time of writing, we would recommend avoiding the use of generated images altogether in culturally-sensitive contexts, especially within the regions with underrepresented cultures to avoid the misrepresentation of cultural identities, cultural inaccuracies, or even potentially harmful stereotypes.

% \dummy{\lipsum[4]}

% \dummy{\lipsum[3]}

% \paragraph{Human-Machine Interaction for a Scalable and High-Quality Annotation Framework}

% \dummy{\lipsum[1]}

\section{Conclusion}
% \todo{Samuel, Onno, Peerat, Joseph}
In this study, we introduced SEA-VL, a corpus-building initiative covering multilingual vision data toward addressing the linguistic and cultural underrepresentation of Southeast Asian languages. By leveraging diverse data collection methodologies---including crowdsourced manual collection, web crawling, and AI-generated images---along with extensive data curation procedures, SEA-VL ensures the creation of high-quality, regionally relevant vision-language datasets that authentically capture the lived experiences of SEA communities.

In summary, our findings highlight the trade-offs in data collection approaches; the potential of web-crawling for producing high-quality, culturally relevant image collections; the limited scalability and maintainability of crowdsourcing; and the limitations of existing image generation models--cultural relevance, naturalness, and licensing issues--for generating accurate, reliable, and scalable culturally relevant images. To promote open-source VL research in SEA, we release our \href{https://huggingface.co/collections/SEACrowd/sea-vl-multicultural-vl-dataset-for-southeast-asia-67cf223d0c341d4ba2b236e7}{SEA-VL dataset} under the CC-BY-SA 4.0 License.
% Through the creation of a rich and diverse SEA cultural dataset, SEA-VL fosters inclusivity, allowing AI systems to more effectively cater to the unique traditions, clothing, architecture, and lifestyles present across SEA. This advancement is crucial in promoting fairness and equality in AI applications, ensuring that models do not disproportionately favor dominant global cultures but instead respect and acknowledge the rich diversity of SEA identities.
% Beyond advancing the representation of SEA cultures from the global perspective, SEA-VL offers an in-depth understanding of trade-offs in data collection approaches through a thorough quality and scalability evaluation. Our findings highlight the potential of producing high-quality culturally relevant image collections through web crawling, the limited scalability and maintainability of crowdsourcing, and the limitations of existing image generation models in terms of performance and legal perspectives, for generating accurate, reliable, and scalable culturally relevant images.

\section*{Limitations}
While SEA-VL is a step forward toward better representations of Southeast Asian culture in multilingual and multimodal AI research, we acknowledge that significant progress is still to be made. We outline several limitations of our study below.

\paragraph{Collection biases and limitations of outreach} Similar to low-resource data collection initiatives such as SEACrowd \cite{lovenia-etal-2024-seacrowd}, CVQA \cite{mogrovejo2024cvqa}, and WorldCuisines \cite{winata2024worldcuisines}, our data collection and outreach practices leveraged mainly on using social media platforms, mailing lists, and internal dissemination through the authors' networks to spread the word about the project. We acknowledge certain limitations in the nature of this practice, including the skewed or imbalanced submissions where countries that are more populous and with better infrastructure, and more connected to the original initiators of the project had higher image contributors (e.g., we saw a substantial higher number of submissions for Indonesia and Singapore compared to Myanmar, Laos, and Cambodia).

Furthermore, collection through self-taken images may only represent more popular cultures being practiced in modern times and require the contributors to be at certain places to take the photo. For future work, an ideal approach would be to have \textit{on-the-ground} representatives for each SEA member country that can assist with data collection across culturally rich areas around SEA (e.g., traveling to rural areas beyond the cities to document non-metropolitan cultural landmarks or food). However, this type of fieldwork requires significant financial resources and manpower. 

\paragraph{Non-holistic representation of deeper, lesser-known culture} Following limitations on data collection, we acknowledge that the cultural representation of SEA is a complex cycle, making it extremely challenging to capture every cultural nuance and requires sustainable and continuous efforts from the community. Hence, our final collected image dataset might not fully represent deeper cultures in SEA at this certain timestep. To address this, we leave the submission portal open beyond the publication of this work to encourage more contributors---especially from underrepresented regions such as Cambodia, Myanmar, Brunei, and Laos---to submit more self-taken culturally relevant images. This will also serve as a good opportunity to conduct better data curation and community involvement as SEA-VL gains wider recognition across SEA.

Moreover, it is important to note that achieving a nuanced cultural curation requires multidisciplinary expertise, which our current approach only partially addresses. For example, images that had strong non-SEA influence because of historical (i.e., colonial legacies) and contemporary (i.e., globalization) reasons were curated in a relatively simplistic and ad-hoc manner. Future research should strive to integrate a systematic guideline from social science experts in order to accurately capture nuances that could have an impact on the resulting models and downstream tasks these datasets will be used for \cite{NEURIPS2024_c07d71ff}.

\paragraph{Potentially limited generalizability using the image dataset} Following certain limitations in our collected image dataset as described in the previous sections, we do not claim that any model trained or optimized using our newly collected culturally relevant SEA image dataset can effectively generalize to emerging cultural practices or underrepresented traditions. We reiterate our plan to make the submission portal open to allow the continued collection of self-taken images from the community to capture said emerging cultural practices and expand the dataset's breadth.

\section*{Ethical Considerations}
We outline several practices we have conducted throughout the project to conform to ethical procedures related to data collection, privacy, and fair attribution.\\
%\todo{Samuel, Onno, Peerat, Joseph}

\noindent \textbf{Responsible credit attribution} We observed justifiable and fair credit practices for our contributors for this project. We draw motivation and guidance from works documenting how low-resource language contributors emphasized lack of recognition (e.g., not being included as a co-author) in past projects related to crowdsourced data collection \cite{ousidhoum2024building}. For image contributors, we used a calibrated pointing system to encourage higher participation from SEA countries with an expected smaller number of active contributors \cite{lovenia-etal-2024-seacrowd} to reach the threshold for co-authorship. The threshold for both image contributors and validators for co-authorship qualification was 200 points. The final arrangement of authors was decided by sorting contributors with the highest garnered points in decreasing order. For more information on the contribution point system used, see Appendix~\ref{app:contribution-point-system}.\\

% Should we insert a table for the pointing system? - Joseph
% Add insights from Nedjma's survey paper here. https://arxiv.org/pdf/2410.12691
% How we properly attribute contributors and how it differs from previous practices (e.g., AYA)

\noindent \textbf{Safety checks for collected images} We performed manual safety checks of the collected image data through consultations with the annotators to ensure that it did not contain sensitive or explicit content (e.g., images with bodily fluids like blood or costumes revealing some private human parts) which may be present in some cultural artifacts from SEA. We instructed annotators to flag and provide additional comments to images within this category for additional review. However, we found that this was not a serious issue as majority of the image submissions were centered on food, landmarks, objects, and everyday life in SEA. \\
% How did we ensure that there are no harmful, unsafe images from the submitted images and from those collected from existing datasets?
% Adding in Joseph's point here from Discord, we'd want to flag religious symbols (images of mosques, Buddhist temples, statues of Buddha); but would want to state that these were captioned in an objective, non-contentious manner -Joel

\noindent \textbf{Censoring personal identifiable information} As part of our submission guidelines, we instructed contributors to remove and blur any personally identifiable information (PII) such as faces, car plates, and house addresses from their images before submitting to the designated form. We recommended using a free third-party PII-remover tool to do this.\footnote{\url{https://picdefacer.com/en/}} Image validators were instructed to flag submissions with non-blurred PII to undergo re-application of the PII remover tool. For any concerns regarding PII in photos, you may contact: \texttt{seacrowd.research@gmail.com}.


\section*{Acknowledgments}
\label{sec:acknowledgments}

We would like to thank our amazing contributors: Srishti Yadav, Raya Ramon, Anwar Choirul Mochammad, Cendekia Airlangga, Wilson Wong, Fernando Julio Cendra, Sabrina Tiun, Derry Wijaya, Randy Zakya Suchrady, Maria Bianca Therese Sta. Monica, Andy Phua, Chernenko Lada, Hendrawan Palgunadi, Dehan Al Kautsar, Elijah J. Gutierrez, Muhammad Razif Rizqullah, Lê Duy Đồng, Hanry Ham, Raymond Ng, Ryan Lau, Atwin Paramudya, Claire, David Samuel, Geoffrey Tyndall, Tuan Anh Vu, Asankhaya Sharma, Febriani Fitria, Pbuakhaw, and Thant Sin Tun for their hard work in submitting and validating cultural image-text pairs for SEA-VL. 

This research is supported by the National Research Foundation, Singapore under its National Large Language Models Funding Initiative. Any opinions, findings and conclusions or recommendations expressed in this material are those of the author(s) and do not reflect the views of National Research Foundation, Singapore. JMI is supported by the National University Philippines and the UKRI Centre for Doctoral Training in Accountable, Responsible, and Transparent AI [EP/S023437/1] of the University of Bath.

\bibliography{custom}

\newpage

\appendix
\onecolumn

\section{Southeast Asian Countries}
\label{app:sea-countries}

\begin{figure}[ht]
    \centering
    \begin{minipage}[b]{0.49\linewidth} % Adjust width for even spacing
        \centering
        \includegraphics[width=\linewidth, trim={0, 0, 0, 0}, clip]{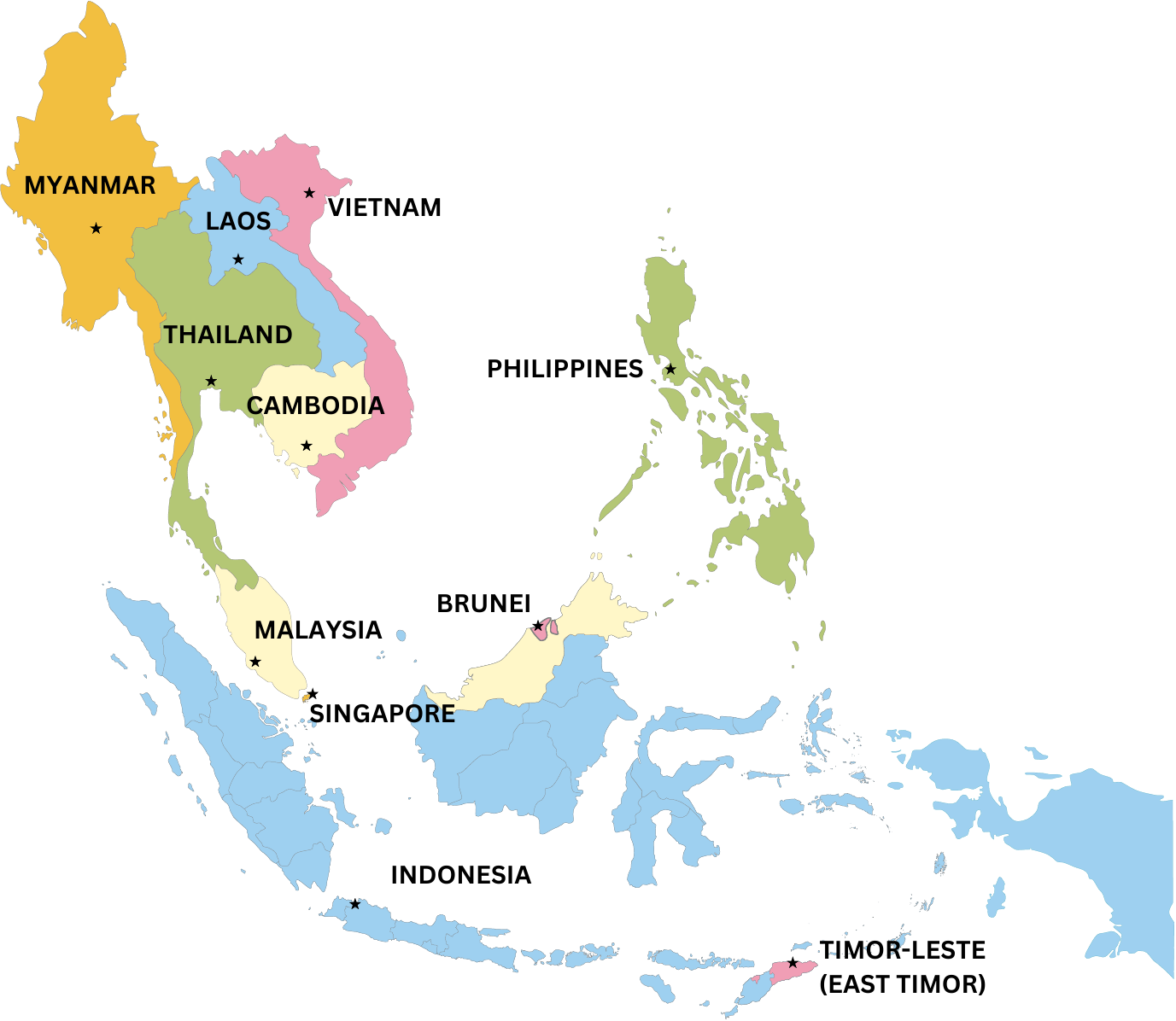}
        \caption{Map of Southeast Asia.}
        \label{fig:sea-map}
    \end{minipage}%
    \hfill
    \begin{minipage}[b]{0.49\linewidth} % Ensure the table fits properly
        \centering
        \resizebox{\linewidth}{!}{
        \begin{tabular}{ccccr}
        \toprule
            \textbf{No.} & \textbf{Abbr.} & \textbf{Country} & \textbf{Flag} & \textbf{Population} \\
        \toprule
            1 & BN & Brunei & \worldflag[width=0.4cm, length=0.8cm]{BN} & 0.5M  \\
            2 & KH & Cambodia & \worldflag[width=0.4cm, length=0.8cm]{KH} & 17.6M \\
            3 & ID & Indonesia & \worldflag[width=0.4cm, length=0.8cm]{ID} & 280.7M  \\
            4 & LA & Laos & \worldflag[width=0.4cm, length=0.8cm]{LA} & 8.0M \\
            5 & MY & Malaysia & \worldflag[width=0.4cm, length=0.8cm]{MY} & 34.6M \\
            6 & MM & Myanmar & \worldflag[width=0.4cm, length=0.8cm]{MM} & 55.8M \\
            7 & PH & Philippines & \worldflag[width=0.4cm, length=0.8cm]{PH} & 114.2M \\
            8 & SG & Singapore & \worldflag[width=0.4cm, length=0.8cm]{SG} & 6.0M  \\
            9 & TH & Thailand & \worldflag[width=0.4cm, length=0.8cm]{TH} & 66.0M  \\
            10 & TL & Timor-Leste & \worldflag[width=0.4cm, length=0.8cm]{TL} & 1.4M  \\ 
            11 & VN & Vietnam & \worldflag[width=0.4cm, length=0.8cm]{VN} & 100.3M \\
            \midrule
            \multicolumn{3}{c}{\textbf{Southeast Asia}} & & \textbf{685.1M} \\
            \multicolumn{3}{c}{Middle East \& North Africa} & & 576.7M \\
            \multicolumn{3}{c}{North America} & & 592.2M \\
            \multicolumn{3}{c}{Europe} & & 774.6M \\
            
        \bottomrule
        \end{tabular}
        }
        \caption{Southeast Asian countries and their populations as of 2025. Efforts and resources in the region still lag behind compared to other, more-represented regions even if SEA's total population is as-large or larger.}
        \label{tab:sea-countries}
    \end{minipage}
\end{figure}

We present an overview of the countries in Southeast Asia (SEA), including their population data in Figure~\ref{tab:sea-countries}, which provides key demographic information for a better understanding of the region's population. We also show a visual representation of SEA region in Figure ~\ref{fig:sea-map}.

\section{Distribution of SEA-VL Dataset}
\label{app:seavl_dist}

% \todo{Ravi, Holy}

\subsection{From Image Crowdsourcing}

% Figure~\ref{fig:seavl_dist}.
% Table~\ref{tab:image-crowdsourcing-accepted-data-stats} presents the statistics of the accepted data constructed through image crowdsourcing.
Table~\ref{tab:image-crowdsourcing-accepted-data-stats} provides an overview of the SEA-VL dataset distribution, detailing the number of accepted images from image crowdsourcing and their cultural relevance scores. A total of 8,018 images were accepted, with an average of 2.6 validators per image. The median relevance score is 4.67, while the average score is 4.38, with a standard deviation of 0.65. Regionally, Indonesia contributes the largest number of images (3,242) with an average relevance score of 4.54. Relevance-wise, it's followed by Cambodia (208), Myanmar (586 images, 4.41), and Malaysia (453 images, 4.38). Other countries such as Thailand, Vietnam, Singapore, and the Philippines are also represented. These statistics highlight the distribution and cultural relevance of images across Southeast Asia.

\begin{table}[t]
    \centering
    \begin{tabular}{lrccc}
        \toprule
        \multicolumn{5}{c}{\textbf{Overall}} \\
        \midrule
        \textbf{\# Data} & 8018 & & \multicolumn{2}{c}{\textbf{Relevance}} \\
        \textbf{\# Validator per data} & 2.6 & & Median & 4.67 \\
         & & & Avg. & 4.38 \\
         & & & Std. & 0.65 \\
        \toprule
        \multicolumn{5}{c}{\textbf{Per region}} \\
        \multicolumn{5}{c}{An image can be relevant for more than one region.} \\
        \midrule
        \textbf{Country} & \textbf{\# Data} & & \multicolumn{2}{c}{\textbf{Avg. relevance}} \\
        Brunei & 72 & & \multicolumn{2}{c}{4.26} \\
        Cambodia & 208 & & \multicolumn{2}{c}{4.54} \\
        Timor-Leste & 12 & & \multicolumn{2}{c}{4.22} \\
        Indonesia & 3242 & & \multicolumn{2}{c}{4.54} \\
        Laos & 157 & & \multicolumn{2}{c}{4.32} \\
        Malaysia & 453 & & \multicolumn{2}{c}{4.38} \\
        Myanmar & 586 & & \multicolumn{2}{c}{4.41} \\
        Phillippines & 543 & & \multicolumn{2}{c}{4.21} \\
        Singapore & 1542 & & \multicolumn{2}{c}{4.07} \\
        Thailand & 1006 & & \multicolumn{2}{c}{4.41} \\
        Vietnam & 541 & & \multicolumn{2}{c}{4.32} \\
        \bottomrule
    \end{tabular}
    \caption{Statistics of accepted data from image crowdsourcing. Relevance refers to image cultural relevance (using 5-point Likert score).}
    \label{tab:image-crowdsourcing-accepted-data-stats}
\end{table}

\begin{figure}
    \centering
    \includegraphics[width=0.775\linewidth]{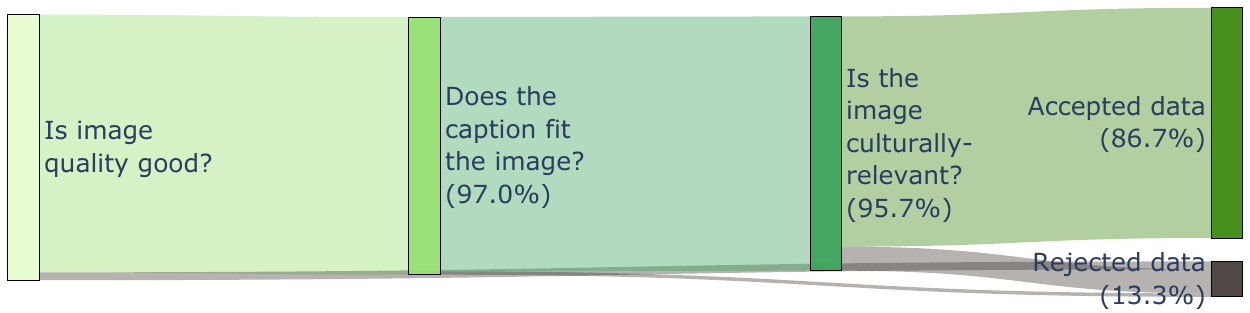}
    \caption{Curating the images obtained from SEA-VL crowdsourcing.}
    \label{fig:image-crowdsourcing-sankey}
\end{figure}

\paragraph{Criteria for Data Quality Flags in SEA-VL Dataset}

To ensure the quality and cultural relevance of the images in the SEA-VL dataset, we define three key evaluation metrics (Figure~\ref{fig:image-crowdsourcing-sankey}).

\begin{itemize}
    \item \textbf{Is the Image Quality Good?}
    \begin{itemize}
        \item \textbf{True}: If the average photo quality score is greater than 0.5 and at least \textbf{two annotators} have reviewed the image.
        \item \textbf{None}: If the average photo quality score is greater than 0.5, but fewer than two annotators provided a review.
        \item \textbf{False}: If the average photo quality score is \textbf{0.5 or lower}.
    \end{itemize}

    \item \textbf{Does the Caption Fit the Image?}
    \begin{itemize}
        \item \textbf{True}: If the average caption fit score is greater than 0.5 and at least \textbf{two annotators} have reviewed the image.
        \item \textbf{None}: If the average caption fit score is greater than 0.5, but fewer than two annotators provided a review.
        \item \textbf{False}: If the average caption fit score is \textbf{0.5 or lower}.
    \end{itemize}

    \item \textbf{Is the Image Culturally Relevant?}
    \begin{itemize}
        \item \textbf{True}: If the average cultural relevance score (found in SEA) is \textbf{3 or higher}, and at least \textbf{two annotators} have reviewed the image.
        \item \textbf{None}: If the average cultural relevance score is \textbf{3 or higher}, but fewer than two annotators provided a review.
        \item \textbf{False}: If the average cultural relevance score is \textbf{below 3}.
    \end{itemize}

    \item \textbf{Overall Data Quality}
    \begin{itemize}
        \item \textbf{True}: If the image meets \textbf{all three} conditions:
        \begin{itemize}
            \item \texttt{average\_photo\_quality} $>$ 0.5
            \item \texttt{average\_found\_in\_SEA\_score} $\geq$ 3
            \item \texttt{average\_caption\_fit} $>$ 0.5
            \item \textbf{At least two annotators} provided reviews
        \end{itemize}
        \item \textbf{None}: If all three conditions are met, but fewer than two annotators reviewed the image.
        \item \textbf{False}: If any of the three criteria do not meet the required threshold.
    \end{itemize}
\end{itemize}

These flags help ensure that the dataset maintains high-quality images, culturally relevant content, and appropriate captions, allowing for more robust research applications.

\subsection{Detailed Statistics of Image Filtering}
\label{app:image-filtering-result}

For the accepted data from image filtering, we begin by filtering three existing datasets. The detailed statistics for all three datasets---Conceptual Captions 3M~\cite{sharma2018conceptual}, COYO (2M)~\cite{kakaobrain2022coyo-700m}, and WiT~\cite{srinivasan2021wit}\footnote{We use the SEA languages subset of WiT from SEACrowd~\cite{lovenia-etal-2024-seacrowd}}---are presented in Table~\ref{tab:small-filtering}, which outlines their alignment with the threshold $\rho$ in our image filtering experiment.

\begin{figure}[!h]
\centering
\begin{minipage}{0.49\textwidth}
\includegraphics[width=\linewidth, trim={2mm 8mm 18mm 0}, clip]{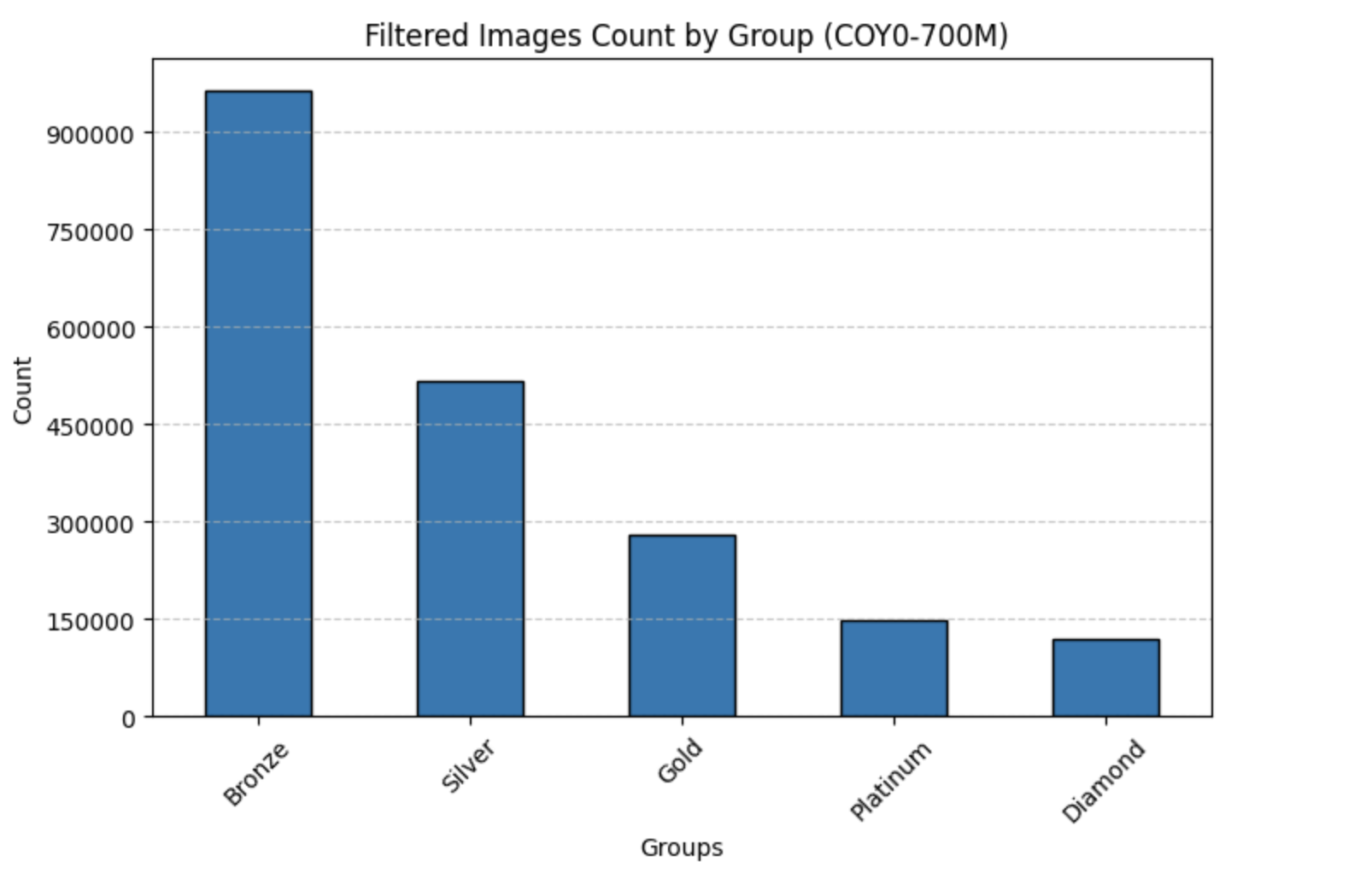}
\end{minipage}
\hfill
\begin{minipage}{0.49\textwidth}
\includegraphics[width=\linewidth, trim={2mm 3mm 2mm 3mm}, clip]{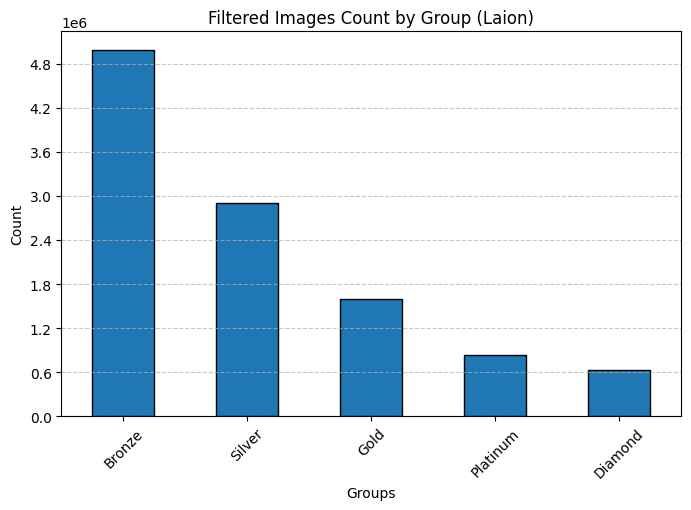}
\end{minipage}
\caption{Histogram of images per filtering threshold in \textbf{(left)} COYO and \textbf{(right)} LAION. The X-axis labels denotes the different threshold groups with Bronze=$[51.5\dots52.5)$, Silver=$[52.5\dots53.5)$, Gold=$[53.5\dots54.5)$, Platinum=$[54.5\dots55.5)$, and Diamond=$\geq55.5$}
\label{fig:scale-filtering}
\end{figure}

To scale up the experiment, we use two large-scale datasets, i.e., COYO (700M)~\cite{kakaobrain2022coyo-700m} and LAION~\cite{schuhmann2021laion}. From 747M image URLS in COYO, we successfully crawled 467.5M images ($\sim$62.5\%), while for LAION we collected 1.3B URLs and gathered 826.5M images ($\sim$66.74\%). We show the histogram for each threshold range for COYO and LAION in Figure~\ref{fig:scale-filtering}.

The accepted data from image filtering is obtained from the Platinum, i.e., $[54.5...55.5)$, and Diamond, i.e., $\geq55.5$ threshold groups in Figure~\ref{fig:scale-filtering}. We then run image deduplication on the combined filtered COYO and LAION data, and end up with a total of 1.28M images.

\section{Additional Detail on Image Crowdsourcing}
\label{app:image-crowdsourcing}

The SEA-VL project page can be accessed at \url{https://seacrowd.github.io/seavl-launch/}.

\subsection{Image Submission}
\label{app:image-crowdsourcing-submit}

\begin{figure*}[!ht]
    \centering
    \includegraphics[width=0.95\linewidth, trim={0, 0, 0, 0}, clip]{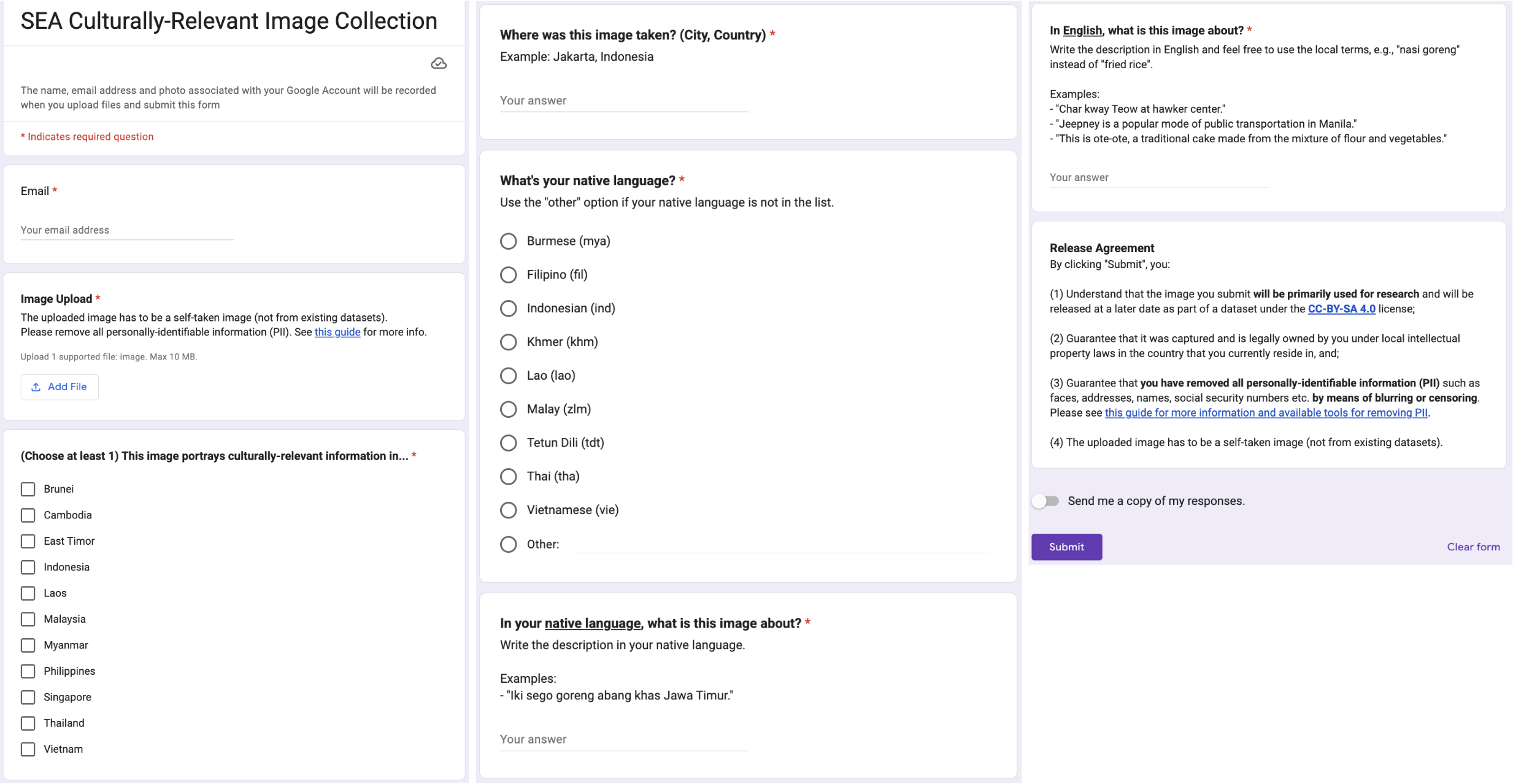}
    \caption{SEA-VL image submission form for single upload.}
    \label{fig:image-submission-form}
\end{figure*}

% \begin{figure*}[!ht]
%     \centering
%     \includegraphics[width=0.375\linewidth, trim={2cm, 2.65cm, 2cm, 1.25cm}, clip]{images/example_image_submission_bulk_uploader.png}
%     \caption{SEA-VL image submission UI tool for bulk upload.}
%     \label{fig:image-submission-bulk-uploader}
% \end{figure*}

% \begin{figure}[!ht]
%     \centering
%     \includegraphics[width=0.55\linewidth, trim={0, 5cm, 2cm, 0}, clip]{images/contrib_progress_image_submission.pdf}
%     \caption{SEA-VL image submission contribution progress.}
%     \label{fig:contrib-image-submission}
% \end{figure}

\begin{figure}[!ht]
    \centering
    \begin{minipage}[b]{0.43\linewidth}
        \centering
        \includegraphics[width=\linewidth, trim={2cm, 2.65cm, 2cm, 1.25cm}, clip]{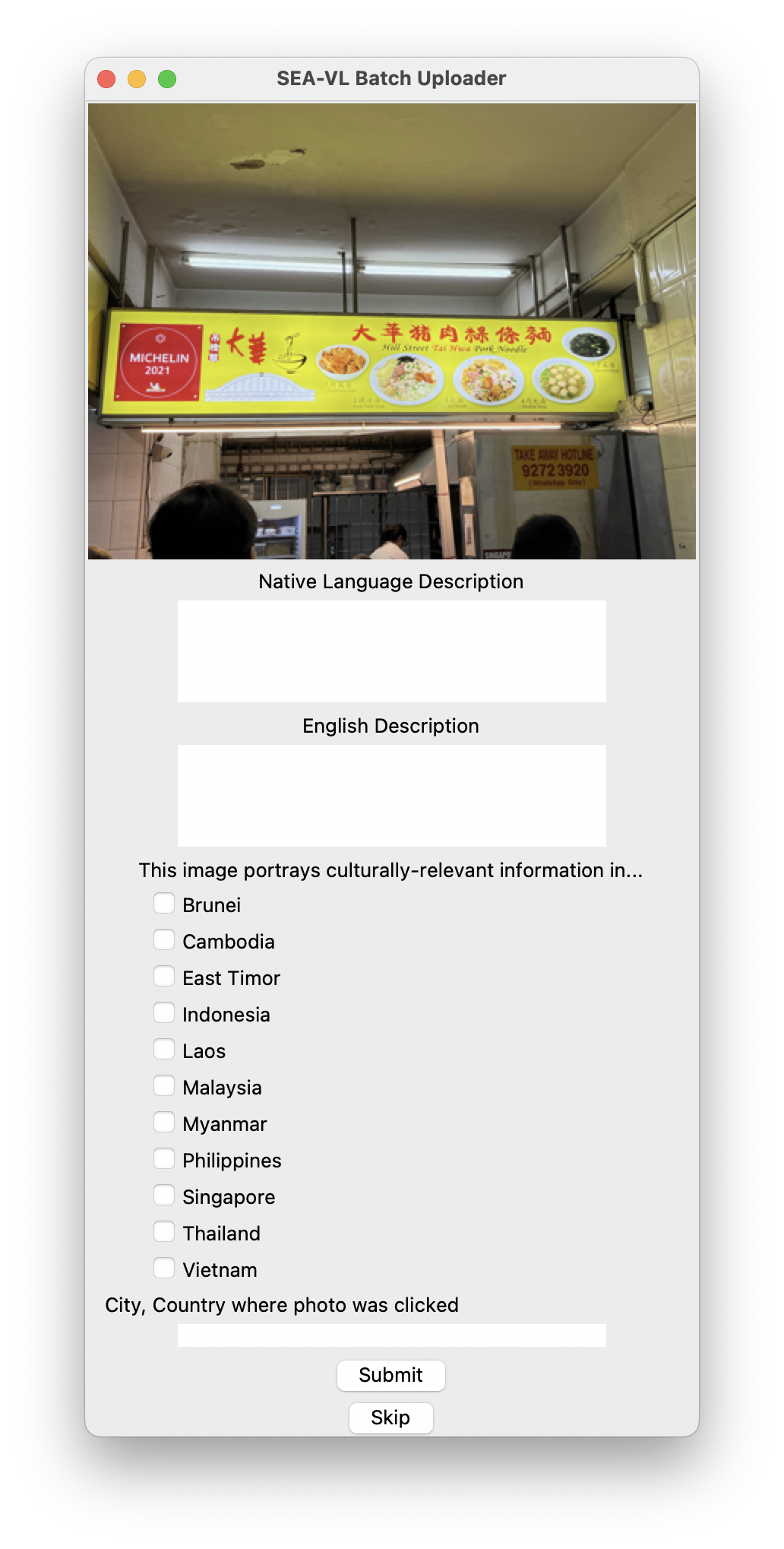}
        \caption{SEA-VL image submission UI tool for bulk upload.}
        \label{fig:image-submission-bulk-uploader}
    \end{minipage}%
    \hfill
    \begin{minipage}[b]{0.55\linewidth} % Ensures even spacing
        \centering
        \includegraphics[width=\linewidth, trim={0, 0, 0, 0}, clip]{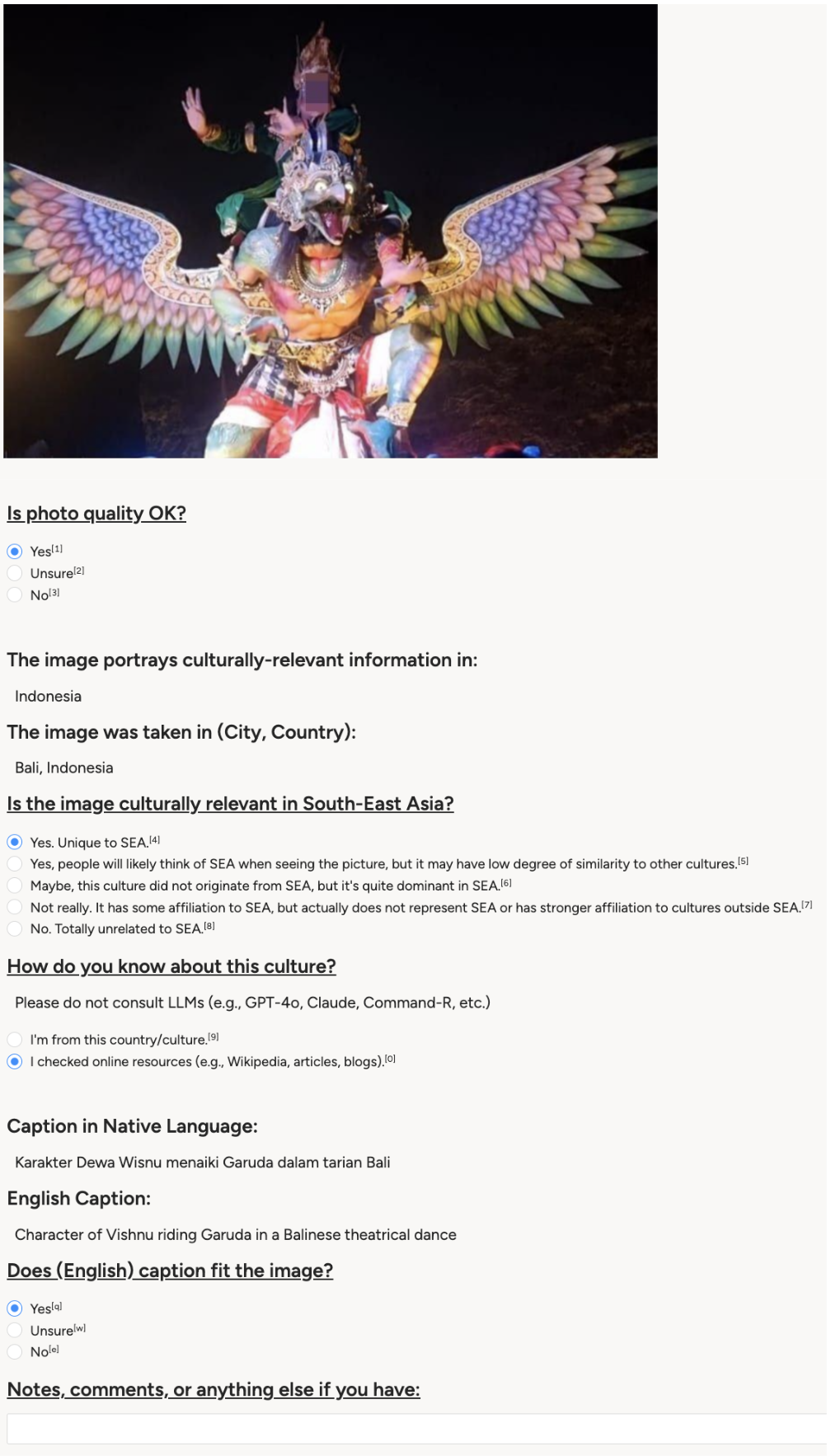}
        \caption{The SEA-VL image validation form used in the quality assurance phase.}
        \label{fig:image-validation-form}
    \end{minipage}%
\end{figure}

The \href{https://docs.google.com/forms/d/e/1FAIpQLScHKqaNlh-SvTD75AtWKkNhFvPNXXDy1eFyrqy3XGXq7M15Vw/viewform}{image submission form} used during the image collection phase is provided in Figure~\ref{fig:image-submission-form}, where contributors input necessary information, including captions and cultural relevance details for each image. Additionally, Figure~\ref{fig:image-submission-bulk-uploader} showcases the bulk upload UI tool, designed to streamline the submission process, allowing contributors to \href{https://github.com/SEACrowd/sea-vl-image-collection/}{upload multiple images at once} while inputting essential metadata. The \href{https://docs.google.com/spreadsheets/d/e/2PACX-1vS5NgqIAxMvN-9tnEbuYab1IP6eM_IFjwKFhp-Le072jiB_zAYvbFA5Be7b0R3RUH3E-anvXIjPa02p/pubhtml?gid=2004297798&single=true}{contribution progress of image submissions} is shown in Figure~\ref{fig:contrib-image-submission}.
% The contribution progress for image submission is illustrated by Figure~\ref{fig:contrib-image-submission}. We provide the image submission form used in the image collection phase in Figure~\ref{fig:image-submission-form}. We provide the screenshot of the bulk upload UI tool used in the image collection phase in Figure~\ref{fig:image-submission-bulk-uploader}.

\subsection{Quality Assurance}
\label{app:image-crowdsourcing-qa}

Contributors validate each submitted image using the form in Figure~\ref{fig:image-validation-form}. Validators assess image quality, ensuring clarity, no offensive content, and that the image is not overly cropped or AI-generated.\footnote{Contributors have to pass a \href{https://forms.gle/p1MmJcKtier9tYZT9}{short screening test} before becoming validators.} Images should also be culturally relevant to Southeast Asia, either by being unique to the region (e.g., local food, landmarks) or strongly reflective of SEA culture (e.g., SEA-specific celebrations). Additionally, validators ensure images do not contain personally identifiable information (PII).

% \begin{figure*}[!ht]
%     \centering
%     \includegraphics[width=0.575\linewidth, trim={0, 0, 0, 0}, clip]{images/example_annotation_form.png}
%     \caption{The SEA-VL image validation form used in the quality assurance phase.}
%     \label{fig:image-validation-form}
% \end{figure*}

% \begin{figure*}[!ht]
%     \centering
%     \includegraphics[width=0.5\linewidth, trim={0, 0, 0, 0}, clip]{images/contrib_progress_image_validation.png}
%     \caption{SEA-VL image validation contribution progress.}
%     \label{fig:contrib-image-validation}
% \end{figure*}

\begin{figure}[!ht]
    \centering
    \begin{minipage}[b]{0.49\linewidth}
        \centering
        \includegraphics[width=\linewidth, trim={0, 5cm, 2cm, 0}, clip]{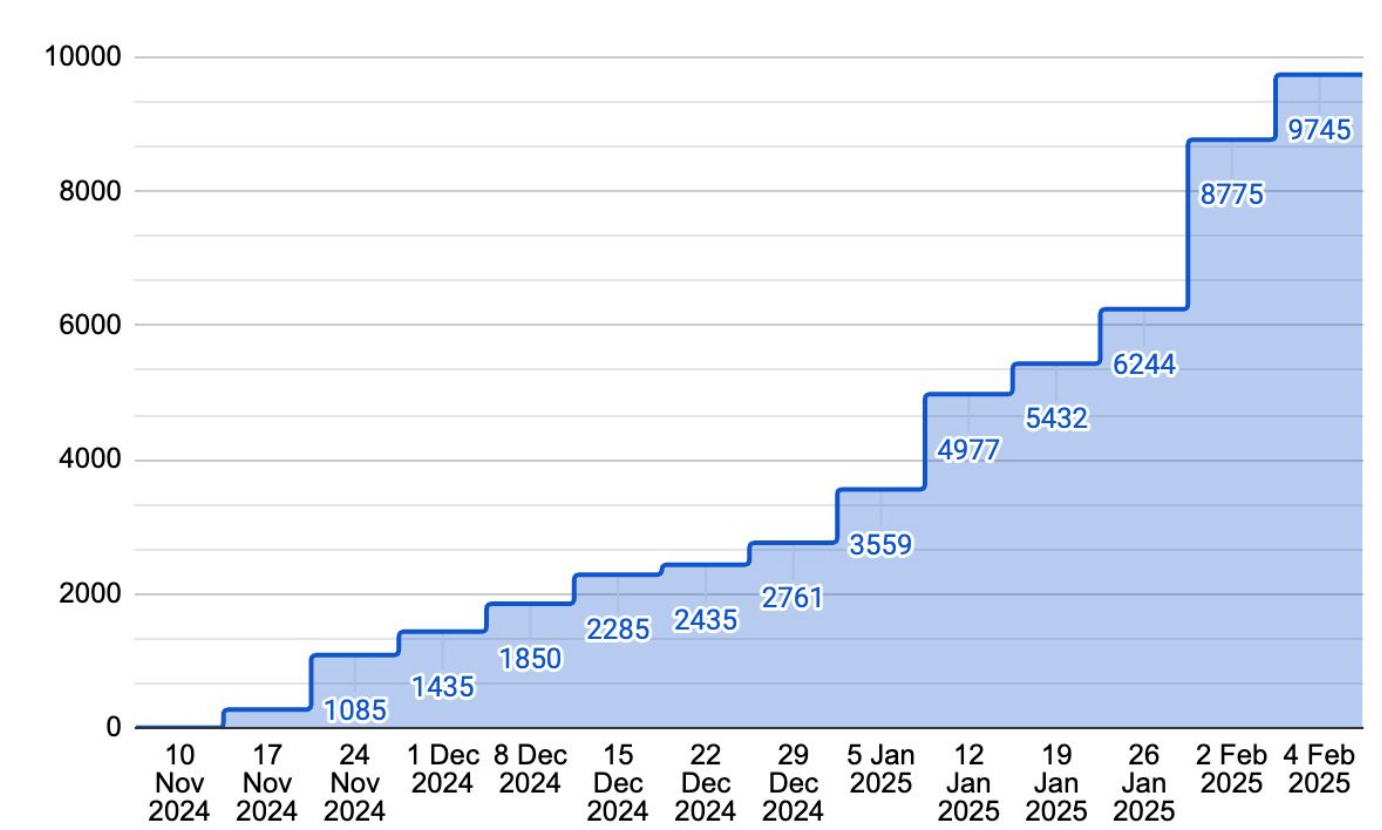}
        \caption{SEA-VL image submission contribution progress.}
        \label{fig:contrib-image-submission}
    \end{minipage}
    \hfill
    \begin{minipage}[b]{0.49\linewidth} % Ensures even spacing
        \centering
        \includegraphics[width=\linewidth, trim={0, 0, 0, 0}, clip]{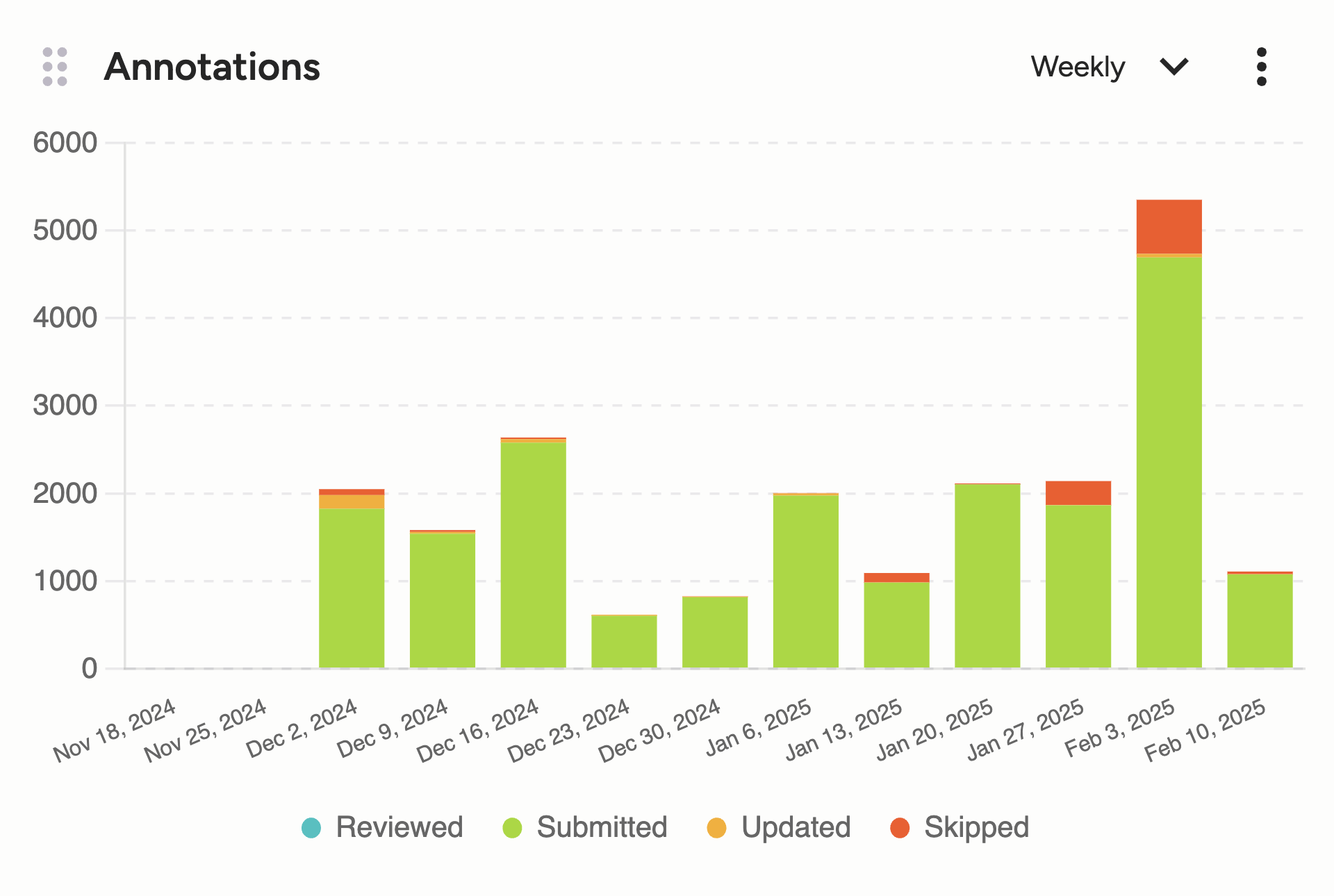}
        \caption{SEA-VL image validation contribution progress.}
        \label{fig:contrib-image-validation}
    \end{minipage}
\end{figure}

\begin{figure*}[!ht]
    \centering
    \begin{tcolorbox}[
        width=\linewidth,
        colback=white, % Background color
        colframe=black, % Border color
        sharp corners,
        boxrule=1pt, % Border thickness
        title={\textbf{SEA-VL Annotation Guideline}}, 
        fonttitle=\bfseries,
        coltitle=white,
        fontupper=\small,
    ]
    
    This document explains the annotation guidelines for SEA-VL data validation. Please note that to become a SEA-VL annotator, you must first pass this short screening test. Once you pass, we will contact you and provide an official validator account to validate our data.

    \textbf{Validation Mechanism}
    
    You will be given an image and its short caption provided by the contributors. We want to ensure that the data is suitable for SEA-VL, that is, relevant to South East Asia and in an acceptable quality. You don’t have to be native in the corresponding SEA countries to validate the data, however, please utilize Google, Wikipedia, or any other trustworthy sources when validating the data that you are not familiar with.

    \textbf{Validation Questions}
    
    Please consult the following guideline for answering the validation questions.\\

    \textbf{1. Is the photo quality OK and appropriate?}  
    
    - Answer OK if the image quality is acceptable. Note that we don’t require high-quality, professional photography. Any smartphone or amateur photography is acceptable.
    Importantly, ensure that the captured object is clear, not overly cropped, and not blurry. Ensure that the photo is also appropriate, e.g., it does not contain harmful stereotypes (eg. High-school gang fight in Indonesia, Tawuran), offensive or suggestive images, or any other illegal content.
    
    - Rotated images are also considered OK, as humans can understand images that are rotated just fine. We want to ensure that our system is robust.
    
    - Note that we expect the images to be originated from the contributors. So if you suspect that the image is taken from the internet or is AI-generated, also select NO.\\

    \textbf{2. Is the image culturally relevant in South-East Asia?}  
    \begin{itemize}
        \item \textbf{Option 1: Yes. Unique to SEA.}
        
        For images or cultures that originate from SEA. Examples include:
        
        - Local food such as Pad Thai.
        
        - Local, unique and very popular buildings like  Petronas Towers or Monumen Nasional.
        
        - Local clothes such as Batik.
        
         - Local activities such as the Ati-Atihan Festival.
        
        - Culturally/historically significant paintings.
        
        - Local brands that are very well-known locally and even internationally (e.g., IKEA is well-known to be Swedish, we want our model to be familiar with our brands too). Eg. Indomie.

        \item \textbf{Option 3: Maybe. Not originally from SEA but very common in SEA culture.}

        For images/concepts not originally from SEA but very ubiquitous in SEA culture and everyday life and/or have SEA-specific local nuances. Examples include:
        
        - Local buildings/places that are not necessarily unique, but have strong SEA-vibe and local influence, e.g., in their architecture. For example, random mosques with local architectures, or random local markets/housing complexes with distinguishable SEA elements. Note that generic homes/hotels that are very typical globally (commonly seen elsewhere outside SEA) should not be considered.
        
        - Non-SEA celebrations with SEA-specific nuances: For instance, Chinese New Year celebrations in Singapore or Eid al-Fitr celebrations in Malaysia.
        
        - Typical concepts or mundane objects with SEA-specific twists: An example could be candy with Rambutan flavor, or objects such as signs, written with SEA languages.
        
        - Foods that are not originated from the region, but very prominent in SEA, e.g. Chinese dishes that are popular in Singapore.
        
        - Non-local brands that are very strong/prominent in SEA (but not generally worldwide), e.g., MSG like Aji no Moto, or Mixue.

        \item \textbf{Option 5: No. Totally unrelated to SEA.}

        Images/concepts that have nothing to do with SEA. Examples include:
        
        - Events like the Super Bowl.
        
        - International landmarks (e.g., Statue of Liberty).
        
        - Generic objects with no cultural relevance (e.g., random chair).
        
        - International brand that has no significance in SEA specifically, eg a photo of a random HSBC office.
        
        \end{itemize}
    
        Select Option 2 (“Yes, people will likely think of SEA when seeing the picture, but it may have a low degree of similarity to other cultures.”) or Option 4 (“Not really. It has some affiliation to SEA, but actually does not represent SEA or has stronger affiliation to cultures outside SEA.”) if you are unsure between corresponding categories.\\
    
        \textbf{3. Does the image contain a person's face, phone number, ID, or car plate numbers?}  
    
        Ensure that the image properly obfuscates personally identifiable information (PII) such as faces, IDs, car plates, etc. Note that the face of a public figure is not considered PII.
        \end{tcolorbox}
    \caption{SEA-VL annotation guideline for data validation.}
    \label{fig:sea-vl-guideline}
\end{figure*}

The \href{https://docs.google.com/document/d/10VTsoD9Lfh_agxE1doSDjgPD4WsDtElDJJ5u775t68g/edit?usp=sharing}{annotation guidelines} (Figure~\ref{fig:sea-vl-guideline}) include 5 options for cultural relevance: Option 1 is for images uniquely associated with SEA, such as local foods or landmarks. Option 2 covers images that strongly reflect SEA culture or lifestyle and have a low degree of similarity to other cultures. Option 3 includes images that may not originally be from SEA but are very common in SEA culture. Option 4 pertains to images with some SEA affiliation but stronger ties to other cultures. Option 5 is for images unrelated to SEA. Regarding the caption, validators must also determine whether it aligns with the image, selecting from three options: yes, no, or unsure (Figure~\ref{fig:image-validation-form}). The contribution progress for image validation is tracked and shown in Figure~\ref{fig:contrib-image-validation}.
% The contribution progress for image validation is illustrated by Figure~\ref{fig:contrib-image-validation}. We provide the image validation form used in the quality assurance phase in Figure~\ref{fig:image-validation-form}. The annotation guideline is presented in Figure~\ref{fig:sea-vl-guideline-page-1} and~\ref{fig:sea-vl-guideline-page-2}.

\newpage
\section{Hyperparameters}
\label{app:hyperparameters}

Our SEA-VL experiment repository can be accessed on GitHub.\footnote{\url{https://github.com/SEACrowd/sea-vl-experiments}}

\subsection{Image Deduplication}

We divided the images into 50 randomly sampled subsets for comparison. Similarity was then assessed using four methods: (1) perceptual hashing via the imagehash\footnote{\url{https://github.com/JohannesBuchner/imagehash}} library (Hamming distance, distance $= 16$, CPU computation); (2) Nomic Embed Vision v1.5 (feature embeddings, threshold $=0.95$, GPU computation); (3) SigLIP (feature embeddings, threshold $=0.85$, GPU computation); and (4) CLIP-ViT (feature embeddings, threshold $=0.95$, GPU computation). These methods differ in their hardware requirements and similarity measurement strategies.

\subsection{Image Generation}

For diffusion models, high-quality images were generated using the hyperparameters specified in the guidance-distilled\footnote{\url{https://huggingface.co/docs/diffusers/main/api/pipelines/flux}} settings of Flux.1-Dev. The inference process was configured with 50 sampling steps and CFG at a scale of 3.5. The scheduler settings remained at their default values, with the Flow Matching Euler Discrete\footnote{\url{https://huggingface.co/docs/diffusers/api/schedulers/flow_match_euler_discrete}} scheduler employed for both Flux.1-Dev and Stable Diffusion 3.5, while the DDIM\footnote{\url{https://huggingface.co/docs/diffusers/api/schedulers/ddim}} scheduler was utilized for Stable Diffusion 2. The generated images have a resolution of \(1024 \times 1024\) pixels. For autoregressive models, we followed the default hyperparameters specified in the official implementation\footnote{\url{https://github.com/deepseek-ai/Janus}}. This configuration included a CFG scale of 5.0, a temperature value of 1.0, and the generation of 576 visual tokens per image. Due to architectural limitations in Janus-Pro, the output images were constrained to a resolution of \(384 \times 384\) pixels.

\subsection{Image Captioning}

\begin{figure}
    \centering
    \includegraphics[width=\linewidth, clip]{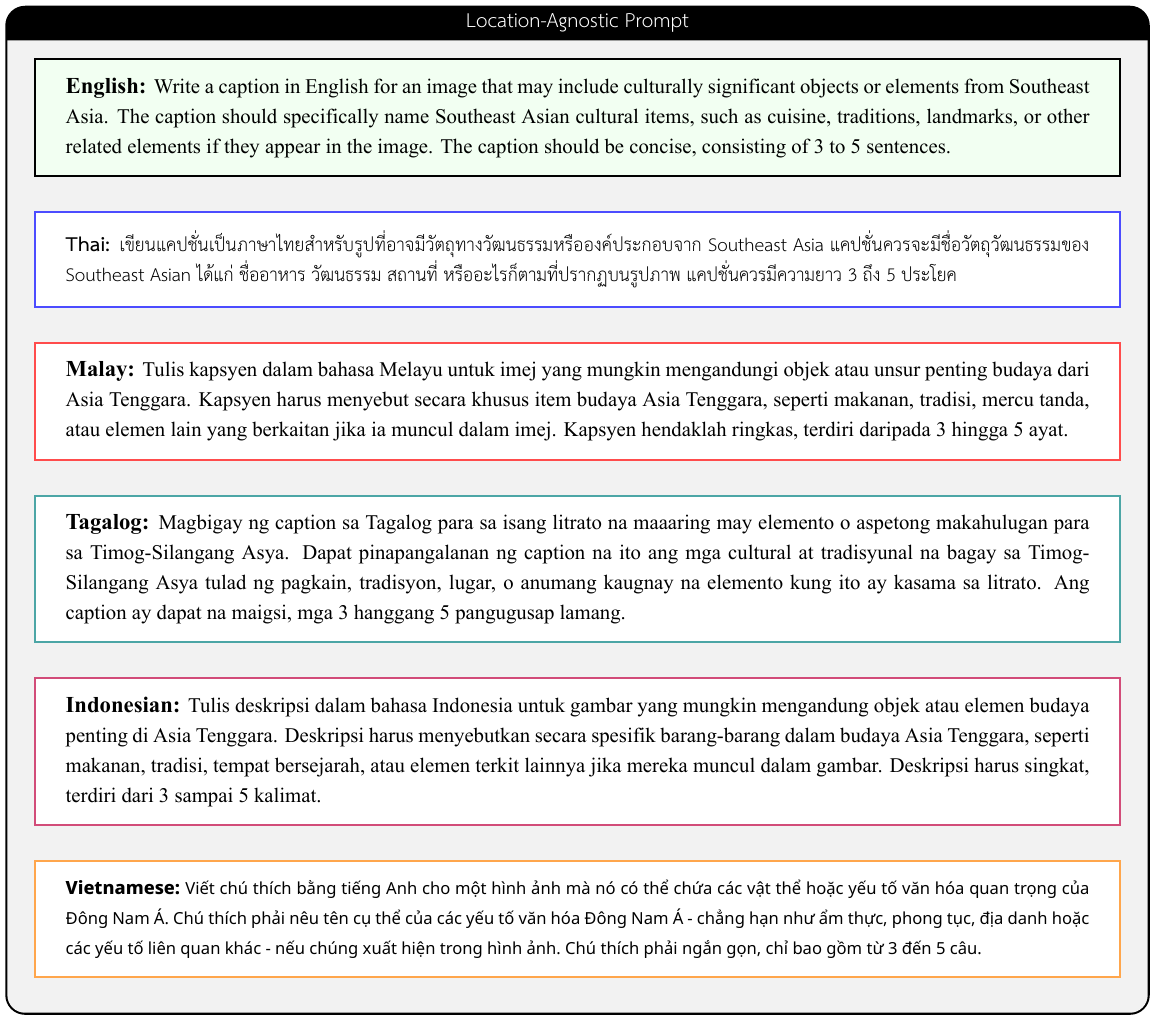}
    \caption{Location-Agnostic Prompts in English and multiple SEA languages.}
    \label{fig:location-agnostic-prompt}
    \vspace{-10pt}
\end{figure}

\begin{figure}
    \centering
    \includegraphics[width=\linewidth, clip]{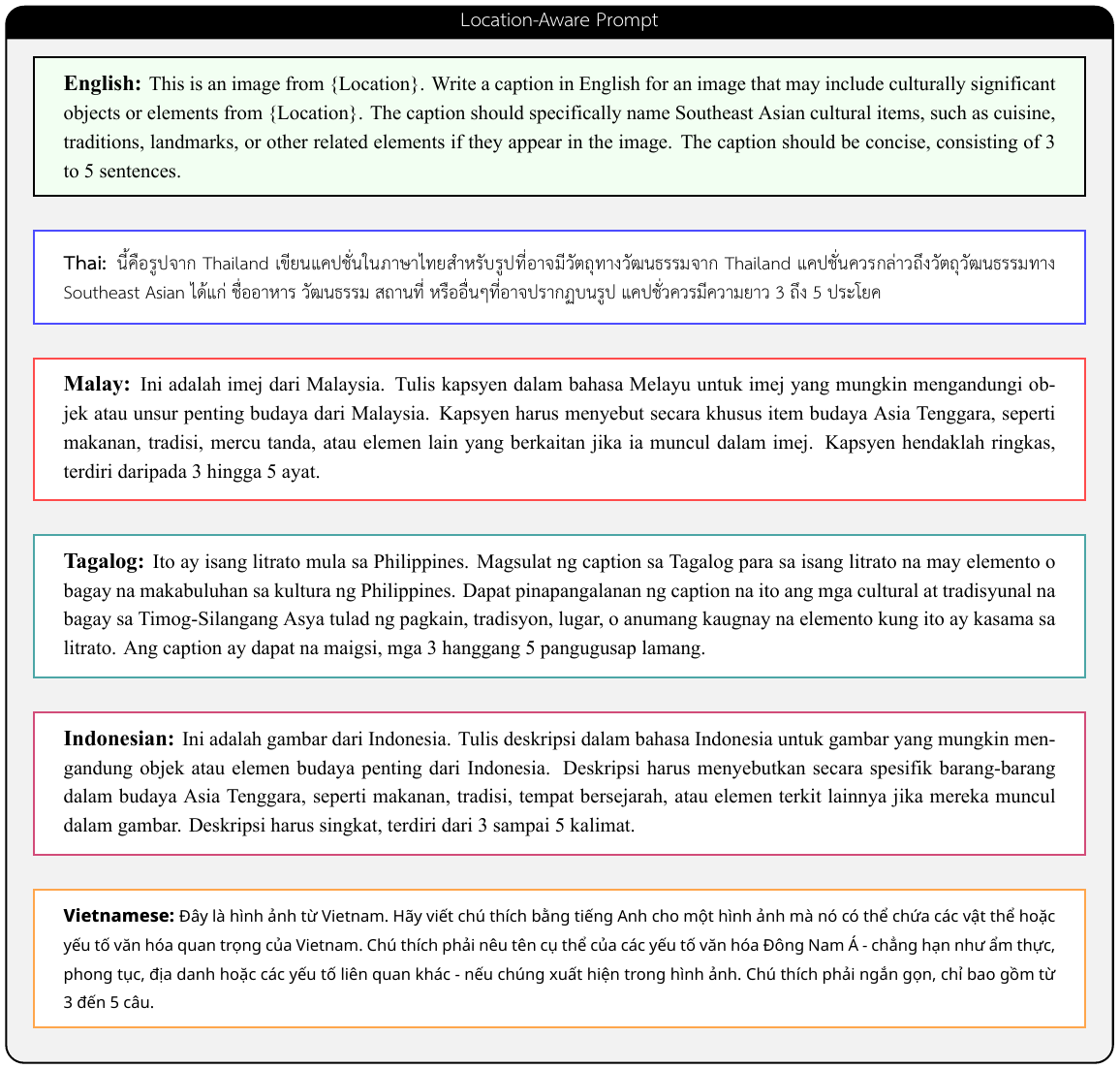}
    \caption{Location-Aware Prompts in English and multiple SEA languages.}
    \label{fig:location-aware-prompt}
    \vspace{-10pt}
\end{figure}

We utilize two types of prompting for image captioning: Location-Agnostic Prompt and Location-Aware Prompt, as detailed in Figure~\ref{fig:location-agnostic-prompt} and~\ref{fig:location-aware-prompt}. The prompt is designed for images that may include culturally significant elements from Southeast Asia. The caption should highlight these cultural items, such as local food, traditions, landmarks, or other relevant elements, and should be concise, consisting of 3 to 5 sentences. In terms of Location-Aware Prompts, the specific country's location is retrieved from the dataset metadata if the image is associated with a specific location within Southeast Asia, and this information is incorporated into the prompt as a hint. In this case, the caption must mention the cultural elements from the specified location, providing more context about the region's culture and traditions.\\

For both prompts, we generate captions in six different languages: English, Thai, Malay, Tagalog, Indonesian, and Vietnamese. This multilingual approach allows for a diverse representation of Southeast Asian culture in different linguistic contexts. Each language provides its unique cultural perspective on the image, ensuring that the captions are both culturally and linguistically appropriate. The captions were generated deterministically using greedy decoding, prioritizing coherence and reproducibility, while disabling the repetition penalty to maintain a fair comparison across languages.

% All captions were generated deterministically using greedy decoding, which prioritizes coherence and reproducibility, while the repetition penalty was disabled for a fair comparison. The prompts used for image captioning in English and multiple SEA languages are presented in Table~\ref{tab:image_captioning_prompts}.

% \newpage
\section{Additional Detail on Image Crawling}

\subsection{Other Methods Explored for Image Filtering}
\label{app:image-filtering-method}

We explore various approaches for image filtering, such as rule-based approaches through URL whitelisting/blacklisting, EXIF geolocation filtering, and other heuristics. Nonetheless, these methods are not very noisy and tend to be source-dependent rendering them unreliable and not scalable. Beyond rule-based approaches, we explore two semantic similarity based approaches, i.e., text-image similarity and image-image similarity. In text-image similarity, we first collect terms that are culturally relevant to SEA, e.g., name of local dishes, name of places, etc. While for image-image similarity, we first collect culturally relevant images from existing source datasets, i.e., CVQA~\cite{mogrovejo2024cvqa} and SEA-VQA~\cite{urailertprasert-etal-2024-sea}.

\begin{table*}[!t]
    \centering
    \resizebox{\linewidth}{!}{
        \begin{tabular}{c|c|c|c|c|c|c|c|c|c}
             \toprule
             \multirow{2}{*}{\textbf{Threshold}} & \multicolumn{3}{c|}{\textbf{CC3M (3M Images)}} & \multicolumn{3}{c|}{\textbf{COYO (1.66M Images)}} & \multicolumn{3}{c}{\textbf{WiT (1.46M Images)}} \\
             \cmidrule(lr){2-4} \cmidrule(lr){5-7} \cmidrule(lr){8-10}
             & \textbf{Relevance} & \textbf{\#Images} & \textbf{\%Images} & \textbf{Relevance} & \textbf{\#Images} & \textbf{\%Images}  & \textbf{Relevance} & \textbf{\#Images} & \textbf{\%Images}  \\
             \midrule
             $<51.5$ & - & 2.99M & 99.12\% & - & 1.65M & 98.58\% & - & 1.44M & 98.25\% \\             $[51.5\dots52.5)$ & 58\% & 11885 & 0.40\% & 54\% & 8925 & 0.54\% & 80\% & 9627 & 0.66\% \\
             $[52.5\dots53.5)$ & 70\% & 6824 & 0.23\% & 40\% & 5919 & 0.36\% & 82\% & 6715 & 0.46\% \\
             $[53.5\dots54.5)$ & 78\% & 3841 & 0.13\% & 70\% & 3996 & 0.24\% & 92\% & 4377 & 0.30\% \\
             $[54.5\dots55.5)$ & 84\% & 2091 & 0.07\% & 82\% & 2323 & 0.14\% & 94\% & 2590 & 0.18\% \\
             $\geq55.5$ & 92\% & 1499 & 0.05\% & 78\% & 2294 & 0.14\% & 90\% & 2162 & 0.15\% \\
             \bottomrule
        \end{tabular}
    }
    \caption{The detailed filtering statistics of the image filtering on CC3M, COYO, and WiT  datasets.}
    \label{tab:small-filtering}
\end{table*}

\subsection{Image Captioning in Local Languages}
\label{app:image-caption-failure}

As discussed in Section \ref{sec:img_captioning}, we initially intended to have image captioning in both English as well as the language corresponding to the respective SEA culture's target language. However, we decided to focus purely on English captions based on insight obtained from an initial pilot study. In this sub-section, we outline the pilot study and describe the findings that justified this choice. 

For each among 5 languages (and their corresponding cultures), we randomly select 10 images and have each image captioned by each of our four chosen multilingual VLMs (Maya (8B)~\cite{alam2024mayainstructionfinetunedmultilingual}, PaliGemma2 (10B)~\cite{steiner2024paligemma2familyversatile}, Pangea (7B)~\cite{yue2024pangeafullyopenmultilingual}, and Qwen2-VL (7B)~\cite{Qwen-VL,Qwen2VL}). These models are prompted with a language-aware prompt, and are instructed to generate captions for the corresponding images in the target language. We thus obtain 200 image-caption pairs. We then manually evaluate the so generated captions on 3 parameters: the correctness of the language the caption was generated in, the correctness of the caption, and the naturalness of the caption. The language correctness is posed as a simple binary yes-no question. The correctness and naturalness of the caption are both measured using a 3-point scale, which we describe in Appendix Section~\ref{app:eval-captioning}.

\begin{figure}[!t]
\centering
\begin{minipage}[b][][t]{0.32\textwidth}
\includegraphics[width=\linewidth]{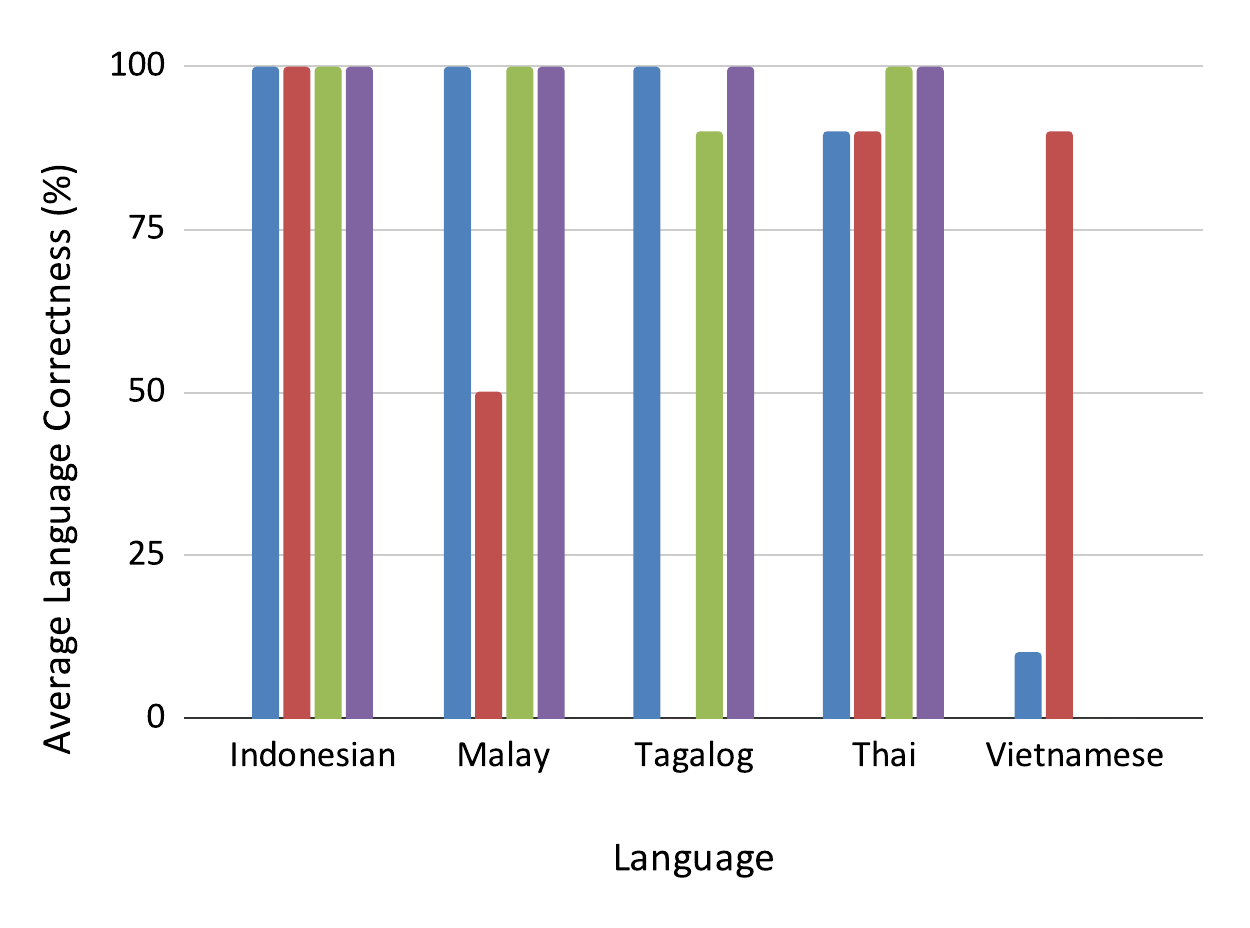}
\end{minipage}
\hfill
\begin{minipage}[b][][t]{0.32\textwidth}
\includegraphics[width=\linewidth, trim={2mm 3mm 2mm 3mm}, clip]{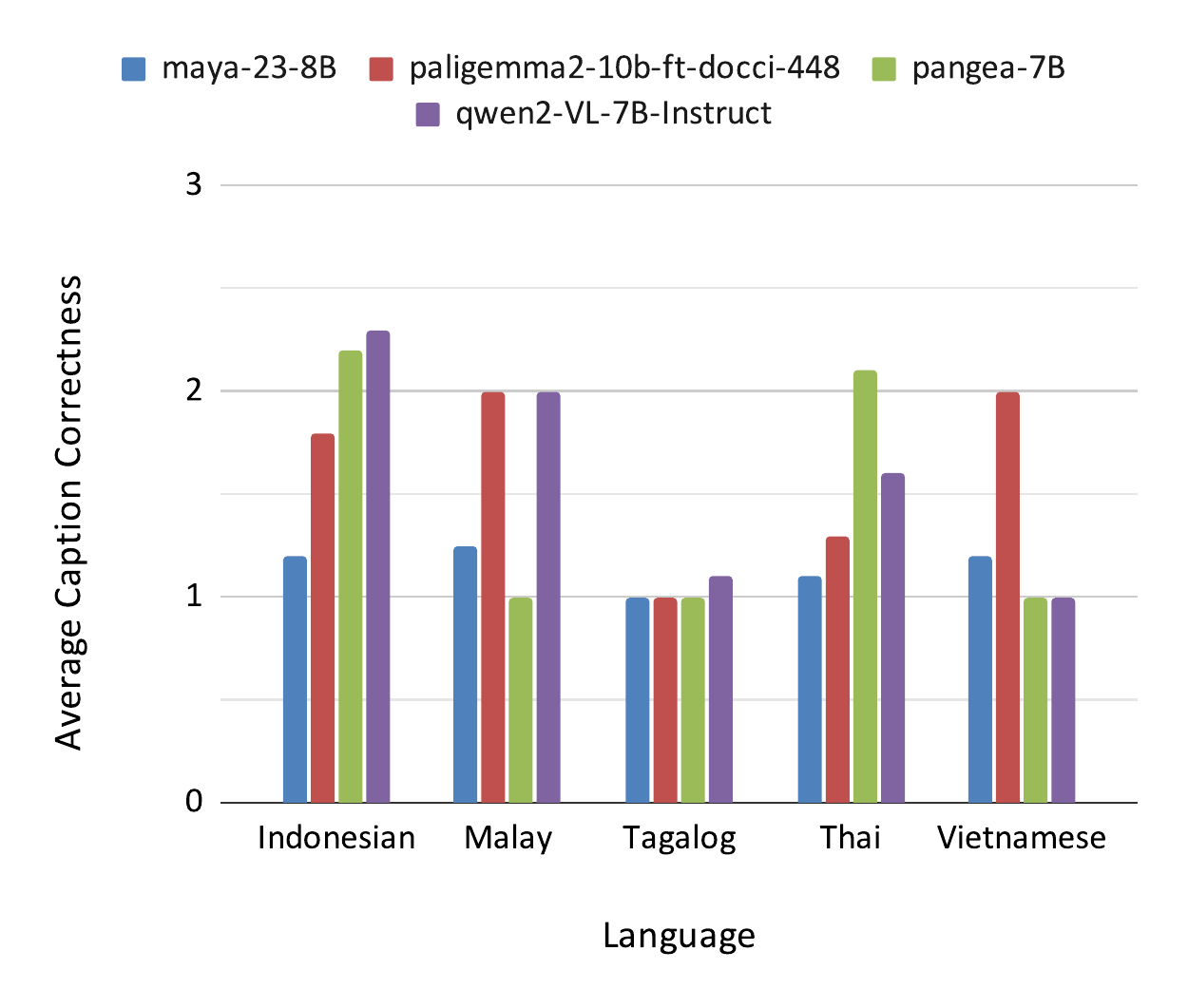}
\end{minipage}
\hfill
\begin{minipage}[b][][t]{0.32\textwidth}
\includegraphics[width=\linewidth, trim={2mm 3mm 2mm 3mm}, clip]{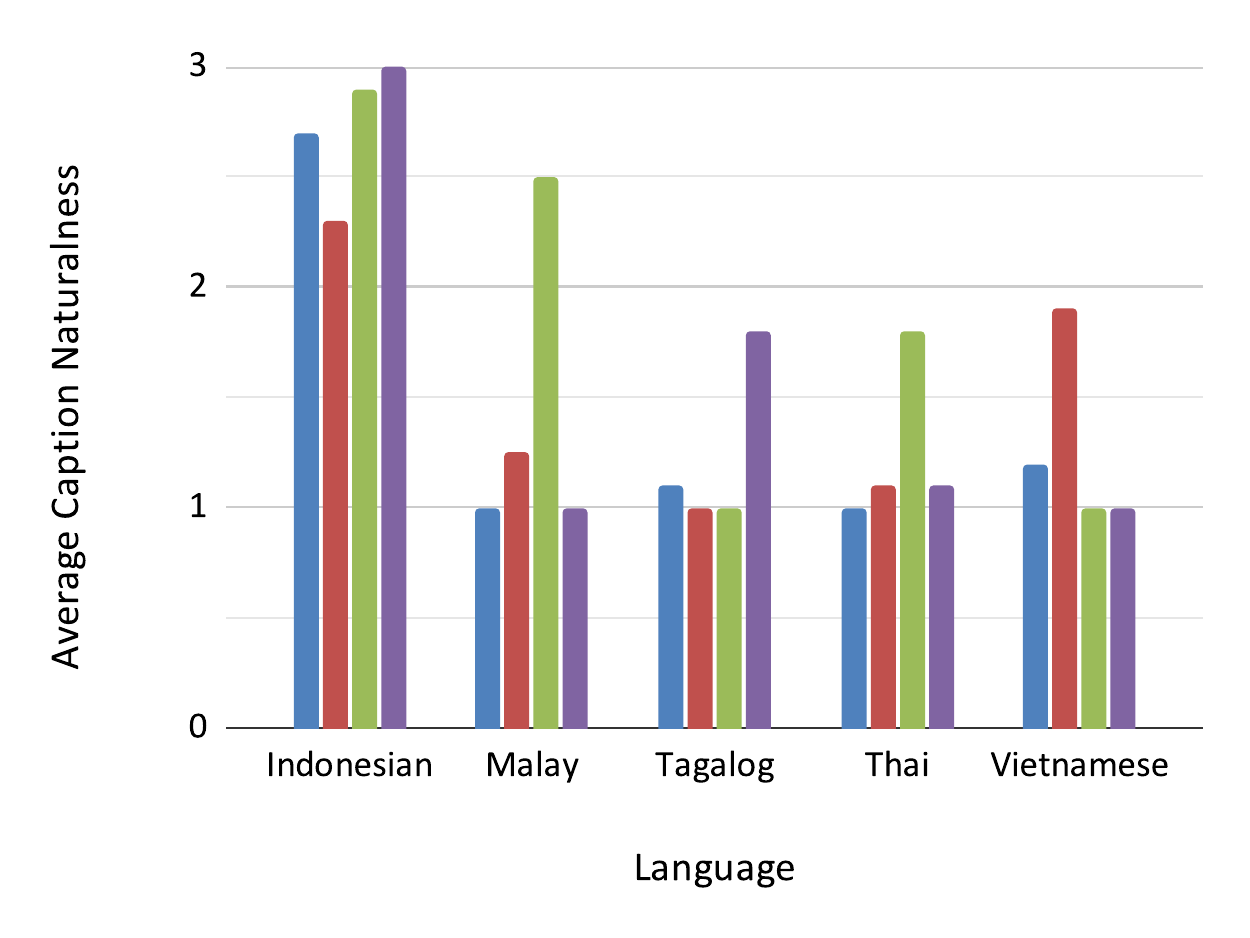}
\end{minipage}
\caption{Bar charts that show performance of various multilingual VLM models on captioning images with a cultural context in the corresponding language, as measured by 3 metrics: \textbf{(left)} the average correctness of the language of the generated caption, \textbf{(center)} the average correctness of the caption, and \textbf{(right)} the average naturalness of the caption.}
\label{fig:native-caption-eval}
\end{figure}

We present our findings in Figure~\ref{fig:native-caption-eval}. Overall, we find that most models struggle to output captions that are correct and natural in the SEA language corresponding to the context of the image shown (the one exception here, perhaps, is Indonesian, where, to our pleasant surprise, we see both Pangea (7B) and
Qwen2-VL (7B) being fairly correct and remarkably natural). Interesting, we find that these models are often unable to respect even the requested language, particularly in the case of Vietnamese; we often find the models defaulting to English captions in these cases.

% in which we manually evaluated captions generated by several LLMs 

% Show pointer to \ref{app:eval-captioning}

% Is there a tie between which languages are lower resource and how LLMs perform?

% Interestingly, this highlights a very interesting opportunity: not only are LLMs poor at naturalness and ..., they can't even stick to the lanague

\section{Contributor Details}
\label{app:author-details}

The details of our authors and their contribution points are provided \href{https://docs.google.com/spreadsheets/d/e/2PACX-1vRPKrLdm6eYRU9g8RBhHHHXZZ8eNXX1KRkvzvNg237GHlbBN6o_d3dgBC4-AGmeZQOXPqETZ6cJ_C-d/pubhtml?gid=1638779097&single=true}{here}. The full contribution point tracking monitor can be accessed \href{https://docs.google.com/spreadsheets/d/e/2PACX-1vSkp3tZJ2LfJp8ajSotMccXQa4SnWZO-cksktMPExH1418to9B4gMB2ds7FMDTZTsHeb8f-Jghdp44o/pubhtml?gid=1432901868&single=true}{here}.

\section{Contributor Demographic}
\label{app:contributor-demographic}

\begin{figure}[!ht]
    \centering
    \begin{subfigure}{0.75\linewidth}
        \centering
        \includegraphics[width=\linewidth]{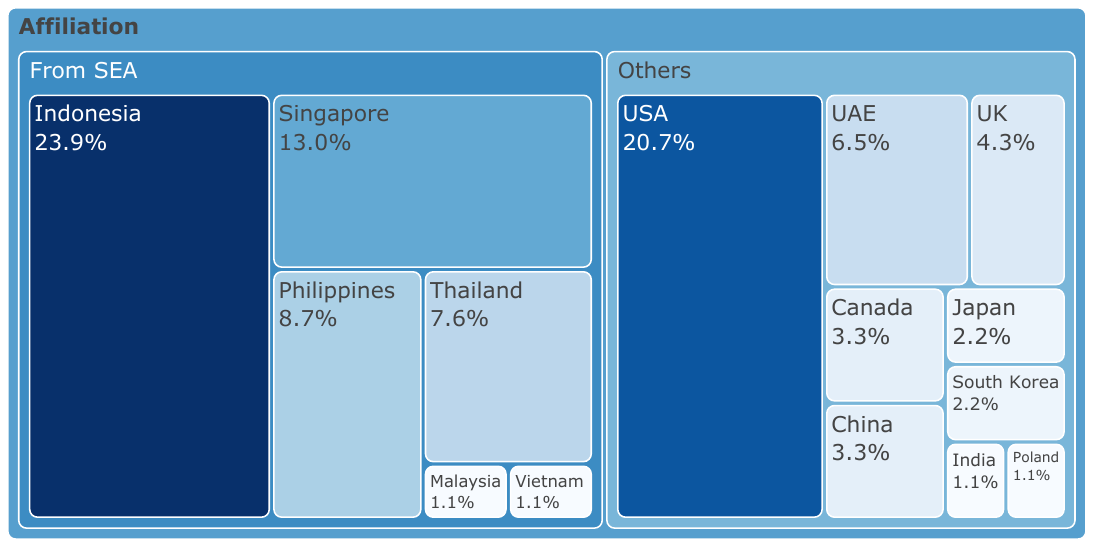}
        \caption{Based on affiliation country}
        \label{fig:author-demographic-affiliation}
    \end{subfigure}
    \begin{subfigure}{0.75\linewidth}
        \centering
        \includegraphics[width=\linewidth]{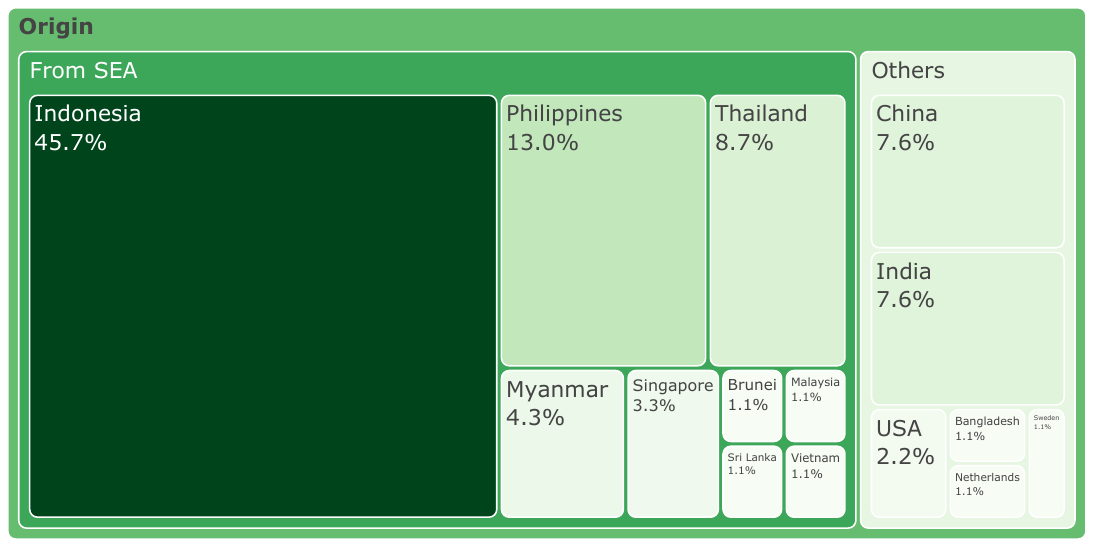}
        \caption{Based on origin country}
        \label{fig:author-demographic-origin}
    \end{subfigure}
    \caption{SEA-VL author demographic based on their affiliation or origin countries.}
    \label{fig:author-demographic}
\end{figure}

\begin{figure}[!ht]
    \centering
    \includegraphics[width=0.65\linewidth]{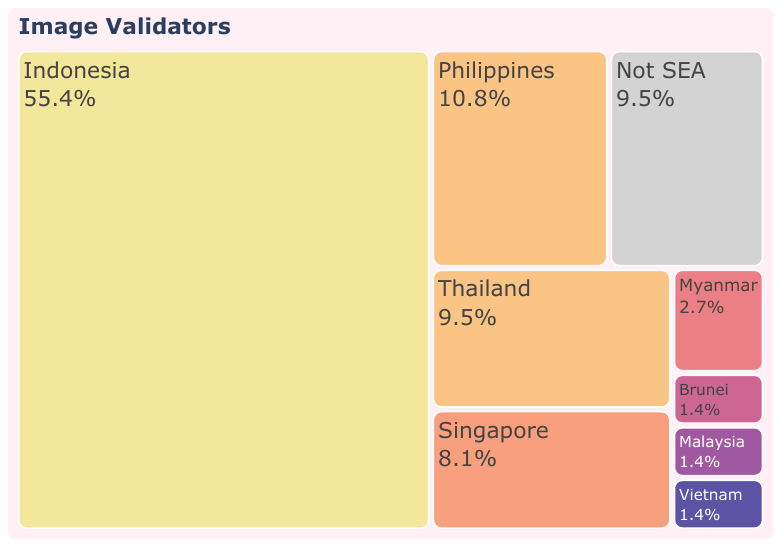}
    \caption{SEA-VL image validator demographic based on their origin countries.}
    \label{fig:validator-demographic}
\end{figure}

We describe our author demographic and image validator demographic in Figure~\ref{fig:author-demographic} and Figure~\ref{fig:validator-demographic}, respectively. Figure~\ref{fig:author-demographic} shows the demographic distribution of SEA-VL authors based on their affiliation and origin countries.

\paragraph{Affiliation (a)} The largest group of authors are affiliated with Indonesia, followed by the USA, Singapore, and the Philippines. Other countries with notable author affiliations include Thailand, the UAE, the UK, and Canada.

\paragraph{Origin (b)} When considering the origin countries, Indonesia again has the largest representation (45.7\%). The Philippines accounts for 13.0\%, followed by Thailand, China, India, and Myanmar. Other countries like the USA, Brunei, and Malaysia are also represented.

In terms of the demographic breakdown of image validators, the majority of image validators are from Indonesia, followed by the Philippines, Thailand, and non-SEA countries. Other countries with smaller representation include Singapore, Myanmar, and Brunei, Malaysia, and Vietnam, as shown in Figure~\ref{fig:validator-demographic}.

\newpage
\section{Contribution Point System}
\label{app:contribution-point-system}

We discuss additional information regarding the contribution point system formulated at the start of the SEA-VL initiative as a form of credit attribution to collaborators from the community. We provide a breakdown of the types of contribution activities and their corresponding points to be awarded in Table~\ref{tab:pointing-system}. For transparency, this point system is discussed at every town hall meeting to inform new collaborators and volunteers to the project. We categorize the types of contributions into two: \textbf{open contribution} which includes image collection and image validation and are open to any potential collaborators and volunteers and; \textbf{closed contribution} which includes performing model experiments, evaluations, paper writing, and management, coordination, and communication of project progress. Tasks under closed contribution are assigned to selected collaborators and original initiators of the project who have the resource and compute capacity such as GPU equipment (see Appendix~\ref{app:hyperparameters} for more details) and experience in paper writing. 

\begin{figure}[!ht]
    \centering
    \includegraphics[width=0.7\linewidth]{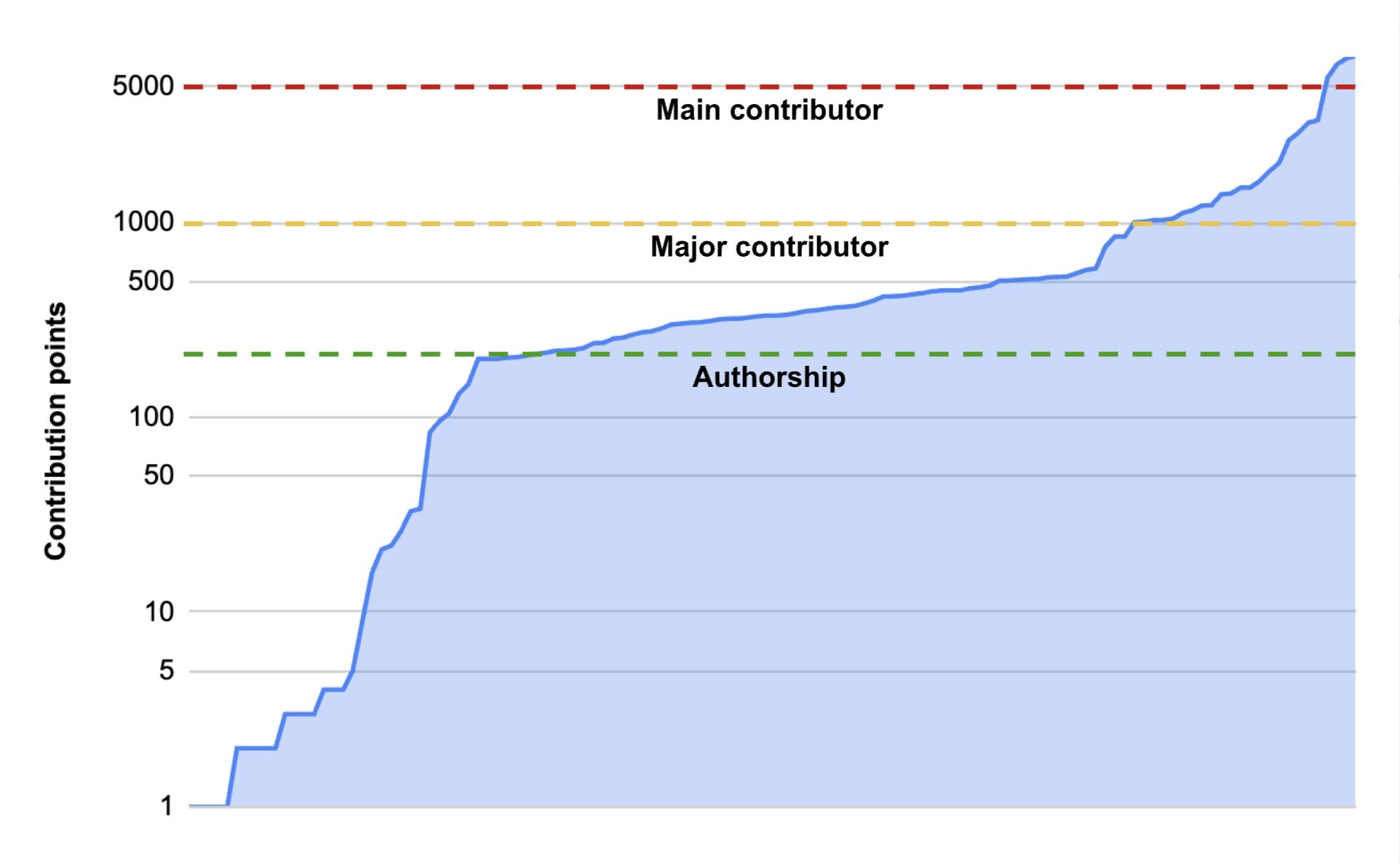}
    \caption{SEA-VL contribution points and thresholds.}
    \label{fig:contribution-points}
\end{figure}

Similar to previous corpus-building initiatives such as SEACrowd \cite{lovenia-etal-2024-seacrowd}, CVQA \cite{mogrovejo2024cvqa}, and WorldCuisines \cite{winata2024worldcuisines}, we set a threshold of \textbf{200 points} for co-authorship denoting significant contribution in this project (Figure~\ref{fig:contribution-points}). We resolve the authorship order based on the decreasing order of points (collaborators with the highest number of points will either come first or come last, depending on their preference). On the other hand, collaborators who did not reach the given threshold will be acknowledged instead.

\begin{table*}[!ht]
\centering
\renewcommand{\arraystretch}{1.2}
\resizebox{0.8\linewidth}{!}{
    \begin{tabular}{@{} >{\raggedright\arraybackslash}p{4cm} >{\raggedright\arraybackslash}p{13cm}}
    \toprule
    \bf Activity & \bf Awarded Points \\ 
    \midrule
    \textbf{Image Collection} & 2 pts per image (Indonesia, Singapore, and the Philippines),  \\&
                                3 pts per image (Thailand, Malaysia, and Vietnam),  \\&
                                4 pts per image (Brunei, Laos, Cambodia, and Myanmar, East Timor) \\
    \midrule
    \textbf{Image Validation} & 1 pt per image \\
    \midrule
    \textbf{Model Experiments} & 100 pts - no limit (based on difficulty, compute resources, and time) \\
    \midrule
    \textbf{Evaluation Procedures} & 100 pts - no limit (based on difficulty, compute resources, and time) \\
    \midrule
    \textbf{Paper Writing} & 100 pts - no limit (based on designated sections) \\
    \midrule
    \textbf{Management, Coordination, and Communication} & 100 pts - no limit (based on difficulty, compute resources, and time) \\
    \bottomrule
    \end{tabular}
}
\caption{The point system used for crediting various forms of contributions from the collaborators of the community. We set \textbf{200 points} as the threshold for co-authorship. We resolve the authorship order based on the decreasing order of points (collaborators with the highest number of points will come first).}
\label{tab:pointing-system}
\end{table*}

\section{Human Evaluation} \label{app:human-eval}

\subsection{Image Filtering Evaluation}
\label{app:eval-filtering}

The Image Filtering Evaluation process involves human annotators assessing images from three datasets: WiT, CC3M, and COYO. All annotators are given the same set of samples for each dataset, where 50 images are randomly selected from each of the following dataset tiers: bronze, silver, gold, platinum, and diamond, resulting in a total of 250 images per dataset. For each image, annotators are asked to classify it as "Yes," "No," or "Not Sure" based on whether the image is relevant to SEA. Three annotators are assigned to WiT and COYO, while five annotators work on CC3M. The evaluation results are then aggregated and averaged across the annotators for each dataset and category. We measure the inter-annotator agreement of the human evaluation using $\gamma$ coefficient~\cite{mathet-etal-2015-unified,mathet-2017-agreement}.

\subsection{Image Duplication Evaluation}
\label{app:eval-duplication}

Given two images, three annotators were asked to assess whether the images are duplicates (binary decision). The metric "duplicated" was defined loosely to the annotators, with annotators encouraged to consider whether having the image pair would be redundant for training purposes. Each annotator was provided with the same set of 75 image pairs, each related to either cuisine or tradition. Finally, for each cultural domain, the duplication score was averaged across the three annotators.

\subsection{Image Generation Evaluation}
\label{app:eval-image-gen}

\begin{table}[h]
\centering
\resizebox{0.85\linewidth}{!}{
\begin{tabular}{p{0.15\linewidth}p{0.75\linewidth}}
\toprule
\textbf{Score} & \textbf{Correctness Description} \\
\midrule
3 & The image correctly describes the given query. \\
2 & The image somewhat correctly describes the given query. \\
1 & The image is irrelevant to the query. \\
\bottomrule
\end{tabular}
}
\caption{Scoring rubric for \textbf{correctness} in \textbf{Image Generation Evaluation}.}
\label{tab:correct-im-gen-eval}
\end{table}

\begin{table}[h]
\centering
\resizebox{0.85\linewidth}{!}{
\begin{tabular}{p{0.15\linewidth}p{0.75\linewidth}}
\toprule
\textbf{Score} & \textbf{Naturalness Description} \\
\midrule
3 & The image is natural and culturally relevant. \\
2 & The image feels somewhat natural. \\
1 & The image is unnatural and looks machine generated. \\
\bottomrule
\end{tabular}
}
\caption{Scoring rubric for \textbf{naturalness} in \textbf{Image Generation Evaluation}.}
\label{tab:natural-im-gen-eval}
\end{table}

Three annotators were assigned to assess the quality of the generated images. Each annotator was tasked with evaluating a distinct set of 250 samples, each focusing on a specific cultural domain: one annotator assessed cuisine, another assessed landmarks, and the third assessed traditions. Each sample consisted of a query (caption) describing an aspect of culture from a specific South East Asian country, along with the corresponding image. The annotators were instructed to evaluate the given caption based on correctness and naturalness according to the rubric in Table~\ref{tab:correct-im-gen-eval} and ~\ref{tab:natural-im-gen-eval} respectively. Finally, for each cultural domain, the correctness and naturalness scores are averaged independently based on the source of the image (whether generated by a model or taken by a human).

\subsection{Image Captioning Evaluation}
\label{app:eval-captioning}

\begin{table}[h]
\centering
\resizebox{0.85\linewidth}{!}{
\begin{tabular}{p{0.15\linewidth}p{0.75\linewidth}}
\toprule
\textbf{Score} & \textbf{Correctness Description} \\
\midrule
3 & The caption correctly describes the given image. \\
2 & The caption somewhat correctly describes the given image. \\
1 & The caption is irrelevant to the image. \\
\bottomrule
\end{tabular}
}
\caption{Scoring rubric for \textbf{correctness} in \textbf{Image Captioning Evaluation}.}
\label{tab:correct-im-cap-eval}
\end{table}

\begin{table}[h]
\centering
\resizebox{0.85\linewidth}{!}{
\begin{tabular}{p{0.15\linewidth}p{0.75\linewidth}}
\toprule
\textbf{Score} & \textbf{Naturalness Description} \\
\midrule
3 & The caption seems to be naturally written by native speakers. \\
2 & The caption feels somewhat natural. \\
1 & The caption is unnatural and looks machine-generated. \\
\bottomrule
\end{tabular}
}
\caption{Scoring rubric for \textbf{naturalness} in \textbf{Image Captioning Evaluation}.}
\label{tab:natural-im-cap-eval}
\end{table}

To assess the quality of generated captions, three annotators were presented with the same set of 450 samples, each consisting of an image along with its corresponding caption. The annotators were instructed to evaluate the given caption based on correctness and naturalness according to the rubric in Table~\ref{tab:correct-im-cap-eval} and ~\ref{tab:natural-im-cap-eval} respectively. Finally, the correctness and naturalness scores are averaged independently based on the source of the generation (whether by a model or a human) across three annotators.

\section{Samples of Images and Captions Collected}

We provide qualitative samples of the collected image-text pairs \href{https://github.com/SEACrowd/seacrowd.github.io/blob/master/docs/SEA_VL_Appendix_J.pdf}{here}.\footnote{\url{https://github.com/SEACrowd/seacrowd.github.io/blob/master/docs/SEA_VL_Appendix_J.pdf}}

\end{document}